\documentclass{imsart}

\RequirePackage{natbib}

\startlocaldefs

\usepackage{amssymb}	
\usepackage{amsthm}		
\usepackage{amsmath}	
\usepackage{mathtools}
\usepackage{enumerate} 
\usepackage{rotate,graphicx,url,color}
\usepackage{subcaption}
\usepackage{soul}

\usepackage{algpseudocode}
\usepackage{algorithm}

\newtheorem{theorem}{Theorem}
\newtheorem{lemma}{Lemma}
\newtheorem{corollary}{Corollary}

\newtheorem{remark}{Remark}[section]
\newtheorem{definition}{Definition}
\newtheorem{assump}{Assumption}

\newcommand{\re}{\mathbb{R}}

\newcommand{\vA}{{\bf A}}

\newcommand{\vC}{{\bf C}}

\newcommand{\vI}{{\bf I}}

\newcommand{\vX}{{\bf X}}

\newcommand{\vx}{\textbf{x}}

\endlocaldefs

\begin{document}

\begin{frontmatter}

\title{Distributed Sparse Linear Regression under Communication Constraints
\support{This research was supported by a grant from the Council for Higher Education Competitive Program for Data Science Research Centers. RF acknowledges the support provided by the Morá Miriam Rozen Gerber Fellowship for Brazilian postdocs, the São Paulo Research Foundation (FAPESP) grant 2023/02538-0, and the JOVEMPESQ grant from UFBA-PRPPG-010/2024.}}
\runtitle{Distributed Sparse Linear Regression}

\author[A]{\fnms{Rodney} \snm{Fonseca}\ead[label=e1]{rodneyfonseca@ufba.br}}
\and
\author[B]{\fnms{Boaz} \snm{Nadler}\ead[label=e2]{boaz.nadler@weizmann.ac.il}}
\address[A]{Department of Statistics, Federal University of Bahia, Salvador-BA, Brazil\\
\printead{e1}}
\address[B]{Department of Computer Science and Applied Mathematics,\\
Weizmann Institute of Science, Rehovot, Israel\\
\printead{e2}}

\runauthor{R. Fonseca and B. Nadler}

\begin{abstract}
In multiple domains, statistical tasks are performed in distributed settings, with data split among several end machines that are connected to a fusion center. 
In various applications, the end machines have limited bandwidth and power, and thus a tight  communication budget. 
In this work we focus on distributed learning 
of a sparse linear regression model, under severe communication constraints. 
We propose several two round distributed schemes, whose communication per machine is sublinear in the data dimension. In our schemes, individual machines compute debiased lasso estimators, but send to the fusion center only very few values. 
On the theoretical front, we analyze one of these schemes and prove that 
with high probability 
it achieves exact support recovery at low signal to noise ratios, where individual machines fail to recover the support.
We show in simulations that our scheme works as well as, and in some cases better, than  more communication intensive approaches. 
\end{abstract}

\begin{keyword}[class=MSC]
\kwd[Primary ]{62J07}
\kwd{62J05}
\kwd[; secondary ]{68W15}
\end{keyword}

\begin{keyword}
\kwd{Divide and conquer}
\kwd{communication-efficient}
\kwd{debiasing}
\kwd{high-dimensional}
\end{keyword}


\end{frontmatter}


\section{Introduction}
    \label{sec:Introduction}

In various applications, datasets are stored in a distributed manner among several sites or machines \citep[chap. 1.2]{fan2020statistical}.
Often, due to communication constraints as well as privacy restrictions, the raw data cannot be shared between machines. 
This motivated the development of methods and supporting theory for distributed learning and inference. See, e.g., the reviews by \citet{huo2019aggregated}, \citet{gao2022review} and references therein.

This paper studies distributed learning of a {\em sparse} linear regression model. 
Consider a response $y\in\mathbb{R}$
and explanatory variable $X\in\mathbb{R}^d$ 
related via
\begin{align}
	y = X^{\top}\theta^* + w,
	\label{e:linear_model}
\end{align}
where $w\sim N(0,\sigma^2)$,   $\sigma>0$ is the noise level, and $\theta^*\in\re^d$.
We focus on a high-dimensional setting $d\gg 1$ and assume that $\theta^*$ is sparse with only $K\ll d$ nonzero coefficients. 
Common tasks
are to 
estimate the unknown vector $\theta^*$ and its support set 
$\mathcal{S}=\{i\in[d]\,|\, |\theta_i^*|>0  \}$. In this work, we consider these tasks under a distributed star-like topology where the observed samples from 
\eqref{e:linear_model} are randomly split among $M$ machines, all connected to 
a fusion center. 

In many distributed regression schemes, each machine sends to the fusion center
its full $d$ dimensional estimate of $\theta^*$, see for example 
\citet{mateos2010distributed, chen2014split, lee2017communication, battey2018distributed,chen2020distributed, liu2021integrative} and references therein.
Hence, these schemes require a communication of at least $O(d)$ bits per machine. 
In some works this is called communication efficient, in the sense that for a machine holding $n$ samples, an $O(d)$ communication is still significantly less than the size {$O(n\cdot d)$} of its data. 

In various contemporary applications, however, 
it is not feasible for end machines to send $O(d)$ bits. This may be due to a constraint on the communication  per machine, or a limit on the total incoming communication at the fusion center. 
An example of the former constraint appears in sensor networks, where multiple devices need to jointly estimate various parameters in real-time and transmit them to a centralized controller. These tasks are performed multiple times and require efficient bandwidth management at each sensor \citep{beuchert2020overcoming, jurado2022survey}.
An example of the latter constraint appears in 
distributed learning in smartphone applications, which often involves millions of machines \citep{bonawitz2019towards}. In this case, even 
if individual machines can send long vectors, the number of machines is so large that the total communication cost at the fusion center may be too high.
Overall, the communication channel is a critical bottleneck in various applications
\citep{kairouz2021advances}. Hence, the design and analysis of communication efficient distributed schemes are important
problems, and even a reduction of one order of magnitude in communication can have a considerable impact.

In this paper, we consider a distributed setting where each machine has a sublinear communication budget
of either $O(d^{1-\alpha})$ bits for some $0<\alpha <1$ or even only $O(K \ln d )$ bits. 
We make the following contributions. On the methodology side, we develop low communication distributed schemes to estimate $\theta^*$ and its support. 
Specifically, in 
Section~\ref{sec:proposed_method} we present a two-round scheme 
whereby the fusion center first estimates the support set $\mathcal{S}$, and next estimates $\theta^*$.
Our scheme has low communication, since in the first round, each machine computes its own debiased lasso estimate, but sends to the center only the indices of its top few largest values.
The fusion center estimates the support of $\theta^*$ by voting, selecting the few indices sent by the largest number of machines. In the second round, the center estimates $\theta^*$ by averaging the least squares solutions on the estimated support set from the first round, as computed by the $M$ machines.

On the theoretical front, in Section \ref{sec:theoretical_results} we derive 
exact support recovery 
guarantees, and mean squared error bounds for our scheme. 
Specifically, Theorem \ref{theorem:support_recovery_large_tau} considers the case where
the communication per machine is extremely limited
to $O(K\ln d)$ bits. Here the signal to noise ratio (SNR) needs to be relatively high. 
Next, Theorem \ref{theorem:support_recovery_thresholds_not_fixed} 
allows for a sub-linear communication of $O(d^{1-\alpha} \ln d)$
and guarantees exact support recovery at lower signal strengths. 
The proofs of our theorems rely on results regarding the distribution of debiased lasso estimators combined with sharp bounds on tails of binomial random variables.

Section~\ref{sec:simulations} presents simulations comparing our schemes to other distributed methods. 
In accordance to our theoretical analysis, these illustrate that in our scheme the fusion center recovers the support $\mathcal{S}$ and consequently accurately estimates $\theta^*$, even at low signal-to-noise ratios where each machine is unable to do so on its own. 
Furthermore, this is achieved with very little communication per machine compared to the dimension $d$.
Interestingly, under a broad range of parameter values, our schemes work as well as, or even better than, more communication-intensive approaches. 
Our simulations also highlight the importance and advantages of a second round of communication. 
Specifically, even though a single-round scheme based on averaging debiased lasso estimates, as proposed by 
\citet{lee2017communication}, is minimax rate optimal and finds the correct support, it nonetheless may output an estimate with a 
significantly 
larger mean squared error than that of our scheme. 
We conclude with a summary and discussion in Section \ref{sec:summary}. 
Proofs appear in the Appendix.

\paragraph{\bf Notations.} 
For an integer $k\geq 1$, we denote $[k]=\{1,2,\ldots,k\}$. The indicator function is denoted as $I(A)$, 
with $I(A)=1$ if 
$A$ holds and zero otherwise. 
The $\ell_q$ norm of a vector $Y\in\re^n$ for {$q\geq 1$} is 
$\|Y\|_q = (\sum_{i=1}^{n} |Y_i|^q)^{1/q}$, 
whereas $\|Y\|_0 = \sum_{i=1}^{n} I(Y_i\neq 0)$ is its number of {nonzero} entries. 
We denote by $|Y|$ the vector whose entries are $(|Y_1|,|Y_2|,\ldots,|Y_n|)$. 
For a $d\times d$ matrix $\vA=\{a_{ij}\}_{i,j=1}^d$, we denote $\| \vA\|_{\infty} = \max_{1\leq i\leq d} \sum_{j=1}^d |a_{ij}|$, whereas  
$\sigma_{\min}(\vA)$ and $\sigma_{\max}(\vA)$ 
denote its  smallest and largest singular values, respectively.
For a subset $J\subset [d]$, $\vA_J$ is the $d\times |J|$ matrix whose columns are those in the subset $J$. Similarly,
$\vA_{J,J}$ is the $|J|\times |J|$ submatrix whose rows and columns correspond to the indices in $J$. 
The cumulative distribution function {(CDF)} of a standard Gaussian is denoted by $\Phi(\cdot)$ whereas $\Phi^c(\cdot)=1-\Phi(\cdot)$. 
We write $a_n \gtrsim b_n$ for two sequences $\{a_n\}_{n\geq1}$ and $\{b_n\}_{n\geq1}$ if there are $C>0$ and $n_0>0$ such that $a_n \geq C b_n$ for all $n>n_0$.



\section{Review of previous works}
\label{sec:background_bibreview}

Distributed schemes for linear regression, not necessarily involving sparsity, have been proposed and theoretically studied in multiple fields, including sensor networks, statistics and machine learning, see for example  \citet{guestrin2004distributed,predd2006distributed, boyd2011distributed,zhang2013communication,heinze2016dual, rosenblatt2016optimality, jordan2019communication,chen2020distributed, dobriban2020wonder, dobriban2021distributed, 
	zhu2021least}.   

\cite{mateos2010distributed} considered distributed {\em sparse} linear regression in a general setting without a fusion center, where all machines communicate with each other. 
They devised a multi-round scheme whereby all the machines reach a consensus and jointly approximate the centralized solution, that would have been computed if all data were available at a single machine. 
Several later works focused on the setting considered in this paper, where $M$ machines are connected in a star topology to a fusion center.
\citet{chen2014split} proposed an approach where each machine 
estimates $\theta^*$ by minimizing a penalized objective with a sparsity-inducing penalty, such as the lasso. 
Each machine sends its sparse estimate to the fusion center, which estimates the support by voting over the indices of the individual estimates of the $M$ machines.
Finally, the center estimates $\theta^*$ by a weighted average of these $M$ sparse estimates.

A limitation of the above approach is that
at each individual machine, its lasso estimate
for $\theta^*$ is biased, and this bias is not attenuated by averaging. 
To overcome this issue, debiased estimators have been applied in various distributed problems, including hypothesis testing, quantile regression, and more \citep{lee2017communication,battey2018distributed,liu2021integrative,lv2022debiased}. 
In particular, \citet{lee2017communication} proposed a single round scheme whereby each machine computes its own debiased lasso estimate and sends it to the fusion center.
The center averages these $M$ vectors and thresholds the result to estimate $\theta^*$ and  
its support. 
\cite{lee2017communication} proved that the resulting estimator achieves the same error rate as the centralized solution, and is minimax rate optimal. 
However, their scheme requires communication of $O(d)$ bits per machine and is thus not applicable in the restricted communication setting considered in this manuscript. 
Moreover, as we demonstrate in the simulation section, unless the signal strength is very low, our two-round scheme achieves a lower mean squared error, with significantly less communication.  
Recently, \cite{amiraz2024recovery} developed a distributed OMP-based method for exact support recovery. Their method, however, requires multiple rounds of communications.

Most related to our paper is the work by \citet{barghi2021distributed}. In their method, each machine computes a debiased lasso estimator $\hat{\theta}$, but sends to the fusion center only indices $i$ for which $|\hat{\theta}_i|$ is above a certain threshold.
The support set estimated by the fusion center consists of all indices that were sent by at least half of the machines, i.e., indices that received at least $M/2$ votes. Focusing on 
consistency of feature selection, \citet{barghi2021distributed} derived bounds on the type-I and type-II errors of the estimated support set. 
In our work, we show both theoretically and empirically, 
that exact support recovery is possible with a much lower voting threshold.
In contrast, with their voting threshold of $M/2$ machines accurate support recovery is possible only at much higher signal strengths.

\section{Background on lasso and debiased lasso}

For completeness, we here briefly review the lasso and debiased lasso
estimators as these are used in our scheme. 
The lasso \citep{tibshirani1996regression} is a 
popular method to fit high-dimensional sparse linear models. 
Given a regularization parameter $\lambda>0$ and $n$ samples $(X_i, y_i)$, stacked in a design matrix $\vX\in\re^{n\times d}$ and a response vector $Y\in\re^n$, the lasso estimator is 
given by 
\begin{align}
	\tilde{\theta} = \tilde{\theta}(\vX, Y,\lambda) 
	= \arg\min_{\theta\in\re^{d}} \left\{ \frac{1}{2n}\|Y - \vX\theta\|_2^2 + \lambda\|\theta\|_1 \right\}.
	\label{e:lasso_estimator}
\end{align}
Several recovery guarantees have been derived for the lasso, assuming the data follows the model (\ref{e:linear_model}) with an exact or approximately sparse $\theta^*$, see  \citep{candes2005decoding,bunea2007sparsity,van2009conditions,hastie2015statistical}.
However, the lasso has two major drawbacks: it may output significantly biased estimates and it does not have a simple asymptotic distribution. 

To overcome these limitations, debiased lasso estimators were derived and theoretically studied \citep{zhang2014confidence,van2014asymptotically,javanmard2018debiasing}. Assuming the rows of $\vX$ have a known population covariance matrix $\Sigma$, \cite{javanmard2014hypothesis} proposed $\frac{1}{n}\Sigma^{-1}\vX^{\top}(Y - \vX\tilde{\theta})$ as a debiasing term.
As $\Sigma$ is often unknown, both \cite{van2014asymptotically} and \cite{javanmard2014confidence} developed methods to estimate its inverse $\Omega=\Sigma^{-1}$. Here, we use the approach of \cite{van2014asymptotically}, 
outlined in Algorithm \ref{alg:decorr_matrix}.
Assuming that $\Omega$ is sparse, 
each column of $\hat\Omega$ is estimated 
by fitting a lasso regression with regularization $\lambda_{\Omega}>0$ to each column of $\vX$ against all other columns. Given $\tilde\theta$ of Eq. (\ref{e:lasso_estimator}) and the matrix $\hat{\Omega}$, the debiased lasso is 
\begin{align}
	\hat{\theta} = \hat{\theta}(Y, \vX,\lambda,\lambda_{\Omega}) = \tilde{\theta} + \frac{1}{n}\hat{\Omega}\vX^{\top}(Y - \vX \tilde{\theta}).
	\label{e:debiased_lasso}
\end{align}
An appealing property of $\hat{\theta}$ is that, under some conditions, it is asymptotically unbiased with a Gaussian distribution \citep{javanmard2018debiasing}.

\begin{algorithm}[t]
	\caption{Computation of a precision matrix estimate $\hat{\Omega}$}\label{alg:decorr_matrix}
	\begin{algorithmic}[1]
		\vspace{0.5em}
		\Require design matrix $\vX\in\re^{n\times d}$, regularization parameter $\lambda_{\Omega}>0$.
		\Ensure precision matrix estimate $\hat{\Omega}\in\re^{d\times d}$.
		\Statex $\vx_i\in \re^{n}$ denotes the $i$-th column of $\vX$.
		\Statex $\vX_{-i}\in \re^{n\times (d-1)}$ denotes the design matrix with the $i$-th column removed.
		\For{$i=1,\ldots,d$}
		\State Fit a lasso with response
		$\vx_i$, design matrix $\vX_{-i}$ and regularization parameter $\lambda_\Omega$.
		\State Let $\tilde{\gamma}_{i}=\{\tilde{\gamma}_{i,j}\}_{j=1,j\neq i}^d\in\re^{d-1}$ be the 
		resulting lasso coefficients from step 2. 
		\State Compute  $\tilde{\tau}_i^2 = (2n)^{-1}\|\vx_i - \vX_{-i}\tilde{\gamma}_{i}\|_2^2 + \lambda_{\Omega}\|\tilde{\gamma}_{i}\|_1$. 
		\EndFor
		\State Construct a $d\times d$ matrix
		\begin{align*}
			\tilde{\vC} = \left[\begin{array}{cccc}
				1 & -\tilde{\gamma}_{1,2} & \cdots & -\tilde{\gamma}_{1,d} \\
				-\tilde{\gamma}_{2,1} & 1 & \cdots & -\tilde{\gamma}_{2,d} \\
				\vdots & \vdots & \ddots & \vdots \\
				-\tilde{\gamma}_{d,1} & -\tilde{\gamma}_{d,2} & \cdots & 1 \\
			\end{array}
			\right].
		\end{align*}
		\State \Return $\hat{\Omega} = \mathrm{diag}\{\tilde{\tau}^{-2}_1,\ldots,\tilde{\tau}^{-2}_d\}\tilde{\vC}$.
	\end{algorithmic}
\end{algorithm}


\section{Distributed sparse regression with restricted communication}
\label{sec:proposed_method}

We consider a setting with $M$ machines connected in a star topology to a fusion center. 
The data $(\vX^{m}, Y^{m})$ at each machine $m$  are $n=N/M$ i.i.d. samples from the model (\ref{e:linear_model}), where $Y^{m}\in\re^n$ and $\vX^{m}\in\re^{n\times d}$. 
The task of the fusion center is to estimate the sparse vector $\theta^*$ of \eqref{e:linear_model}, under a constraint of limited communication with each machine. 
We present a two-round scheme for this task. For simplicity, we assume that the noise level $\sigma$ is known. If $\sigma$ is unknown, it may be consistently estimated by the scaled lasso  \citep{sun2012scaled},  see also  \cite[Corollary 3.10]{javanmard2018debiasing}.
{We remark that neither the fusion center nor the individual machines need to know the sparsity $K$.}

\begin{algorithm}[t]
	\caption{Distributed voting based scheme for support estimation}\label{alg:count_votes}
	\begin{algorithmic}[1]
		\vspace{1em}
		\Require Data $(\vX^m, Y^m)\in\re^{n\times(d+1)}$, vote threshold $V_T>0$, coefficient threshold $\tau>0$ and regularizations $\lambda_{\Omega}, \lambda >0$.
		\Ensure Support estimate $\hat{\mathcal{S}} \subseteq [d]$.
		\smallskip
		\Statex \textbf{At each local machine $m=1,\ldots,M$}
		\State Compute a lasso estimator $\tilde{\theta}^m$ via Eq. (\ref{e:lasso_estimator}) with regularization parameter $\lambda$.
		\State Compute a precision matrix estimate $\hat{\Omega}^m\in\re^{d\times d}$ by  Algorithm \ref{alg:decorr_matrix} with $\vX^m$ and $\lambda_{\Omega}$.
		\State Compute a debiased lasso estimate $\hat{\theta}^m\in\re^{d}$, Eq.  \eqref{e:debiased_lasso}, with data {$(\vX^m, Y^m)$}, $\lambda$ and $\hat{\Omega}^m$.
		\State Calculate the empirical covariance matrix $\hat{\Sigma}^{m}=n^{-1}(\vX^m)^{\top} \vX^m$.
		\State 
		Use $\hat{\Omega}^m$ and $\hat{\Sigma}^{m}$ to compute the 
		normalized vector $\hat{\xi}^{m}\in\re^d$ of Eq. \eqref{e:standardized_debiased_lasso}.
		\State Set $\mathcal S^{m} = \big\{i \in [d] \,\big|\, |\hat{\xi}^{m}_{i}| > \tau \big\}$ and send it to the fusion center.
		\smallskip
		\Statex \textbf{At the fusion center}
		\State For each $i\in[d]$, compute $V_i = \sum_{m=1}^M  I(i \in \mathcal S^m)$.
		\State \Return $\hat{\mathcal{S}}=\{i \in [d] \,|\, V_i > V_T \}$.
	\end{algorithmic}
\end{algorithm}

The first round of our scheme is outlined in 
Algorithm \ref{alg:count_votes}, whereas Algorithm~\ref{alg:two_round_scheme} outlines the full two-round scheme. 
In the first round, each machine $m\in[M]$ computes the following quantities
using its data $(\vX^m, Y^m)$: 
(i)~a lasso estimate $\tilde{\theta}^m$ by Eq. \eqref{e:lasso_estimator}; (ii) a  matrix $\hat{\Omega}^m$ by Algorithm \ref{alg:decorr_matrix}; and (iii) a debiased lasso $\hat{\theta}^m$ by Eq. \eqref{e:debiased_lasso}.
Up to this point, this is identical to  \citet{lee2017communication}. 
The main difference is that in their scheme, each machine sends to the center its debiased lasso estimate $\hat{\theta}^m \in \mathbb{R}^d$, incurring $O(d)$ bits of communication per machine. 
In contrast, in our scheme, the goal of the first round is only to estimate the support of $\theta^*$ and not $\theta^*$ itself. 
To this end, after steps (i)-(iii) above, each machine computes the following normalized vector $\hat{\xi}^m$,
\begin{align}
    \hat{\xi}_k^{m} = \frac{\sqrt{n}\hat{\theta}^{m}_k}{\sigma(\hat{\Omega}^m\hat{\Sigma}^{m}(\hat{\Omega}^m)^{\top})_{kk}^{1/2}}
        , 
    \qquad  \forall k\in[d].
\label{e:standardized_debiased_lasso}
\end{align}
Each machine sends to the center only the indices $k$ 
that satisfy $|\hat{\xi}_k^{m}|>\tau$ for a suitable threshold $\tau>0$.
For $\tau$ high enough, the number of sent indices is much lower than $d$. 
Given the messages sent by the $M$ machines, the fusion center 
estimates the support set $\hat{\mathcal S}$ by voting, taking only those indices that were
sent by at least $V_T$ machines, for a suitable voting threshold $V_T>0$. 

In the second round, the fusion center sends the set $\hat{\mathcal{S}}$ to all $M$ machines. 
Next, each machine 
computes the least squares solution, restricted to the set $\hat{\cal{S}}$, \begin{equation}
    \hat\beta^{m} = \arg\min_{\beta}
        \| \vX_{\hat{\mathcal{S}}}^m \beta - Y^m\|_2^2, \qquad  \forall m\in[M],
            \label{e:beta_m}
    \end{equation}  
where $\vX_{\hat{\mathcal{S}}}^m \in \mathbb{R}^{n\times |\hat{\cal S}|}$ consists of the columns of $\vX^m$ corresponding to the indices in $\hat{\cal S}$. 
Each machine then sends its vector $\hat \beta^m$ to the fusion center. 
Finally, the center estimates $\theta^*$ by averaging these $M$ vectors,  
\begin{equation}
    \hat \theta_i = 
    \left\{
        \begin{array}{cc}
             \frac1M \sum_{m=1}^M \hat \beta^m_i &  i \in \hat{\cal S}, \\
             0  & i\not\in \hat{\cal S}.
        \end{array}
    \right. 
            \label{e:theta_hat}
\end{equation}

\begin{algorithm}[t]
\caption{Two round distributed scheme to estimate $\theta^*$}\label{alg:two_round_scheme}
\begin{algorithmic}[1]
\vspace{1em}
\Require Data $(\vX^m, Y^m)\in\re^{n\times(d+1)}$ from each machine $m\in[M]$, vote threshold $V_T>0$, coefficient threshold $\tau>0$ and regularization parameters $\lambda_{\Omega}, \lambda >0$.
\Ensure A two-round estimate $\hat{\theta}\in\re^d$ of $\theta^*$ with support $\hat{\mathcal{S}} \subset [d]$.
\smallskip
\Statex \underline{First round}
\smallskip
\State The fusion center uses $V_T$, $\tau$ and the regularization parameters to compute $\hat{\mathcal{S}}$ with Algorithm \ref{alg:count_votes} (voting scheme) or Algorithm \ref{alg:count_signed_votes} (sign-based scheme).
\smallskip
\Statex \underline{Second round}
\smallskip
\State The fusion center sends $\hat{\mathcal{S}}$ to all $M$ machines.  
\smallskip
\Statex \textbf{At each local machine $m=1,\ldots,M$}
\State Compute $\hat{\beta}^m = 
 \arg\min_\beta \| \vX_{\hat{\mathcal{S}}}^m \beta - Y^m  \|_2^2$.
\State Send $\hat{\beta}^m$ to the fusion center.
\smallskip
\Statex \textbf{At the fusion center}
\State Given $\hat\beta^1,\ldots, \hat\beta^M$, compute the estimate 
$\hat\theta \in \re^d$ via Eq. \eqref{e:theta_hat}.
\State \Return $\hat{\theta}$.
\end{algorithmic}
\end{algorithm}

\begin{remark}
The communication of the first round 
depends on the threshold $\tau$. A high threshold 
leads to only a few sent indices. 
However, at low signal strengths, the support indices $k\in \mathcal S$ may not satisfy $|\hat{\xi}^m_k|>\tau$ and thus may not be sent. 
A lower threshold $\tau$ allows support recovery of weaker signals, but
at the expense of many more indices sent by each machine. 
Since the maxima of $d$ standard Gaussian variables scales as $\sqrt{2\ln d}$, to comply with the communication constraints, the threshold $\tau$ should also scale as $O(\sqrt{\ln d})$. 
In Section \ref{sec:theoretical_results}, we present suitable thresholds and 
sufficient conditions on the number of machines and on the signal strength, which guarantee support recovery by Algorithm \ref{alg:count_votes}, with high probability and little communication per machine.
\end{remark}

\begin{remark}
	In the second round each machine sends $|\hat{\cal S}|$ values. At the expense of sending $O(|\hat{\cal S}|^2)$ values, 
    the fusion center can compute the {exact} centralized least squares solution corresponding to the set $\hat{\cal S}$, denoted $
    \hat\theta^{\mbox{\tiny LS}}$. 
    Specifically, suppose that each machine sends to the center 
    both   the vector 
    $ (\vX_{\hat{\mathcal{S}}}^m)^\top Y^m$
    of length $|\hat{\cal S}|$, and the 
    $|\hat{\cal S}| \times |\hat{\cal S}|$
    matrix $ (\vX_{\hat{\mathcal{S}}}^m)^\top 
    \vX_{\hat{\mathcal{S}}}^m$. 
    The center can compute $\hat\theta^{\mbox{\tiny LS}}$
as follows, 
    \begin{equation}
        \hat\theta^{\mbox{\tiny LS}} = \left(
        \sum_{m=1}^M 
        (\vX_{\hat{\mathcal{S}}}^m)^\top 
    \vX_{\hat{\mathcal{S}}}^m
        \right)^{-1}
        \sum_{m=1}^M (\vX_{\hat{\mathcal{S}}}^m)^\top Y^m .
        \label{e:centralized_ols_estimator}
    \end{equation}
    This modification is relevant only
    if $|\hat{\cal S}|^2 \ll d$.      
    For $|\hat{\cal S}| \propto \sqrt{d}$, this scheme incurs communication linear in $d$, comparable to sending the full debiased lasso vector. 
    \end{remark}

\begin{algorithm}[t]
	\caption{Sign-based distributed support estimation scheme}\label{alg:count_signed_votes}
	\begin{algorithmic}[1]
		\vspace{1em}
		\Require Data $(\vX^m, Y^m)\in\re^{n\times(d+1)}$ from each machine $m\in[M]$, vote threshold $V_T>0$, coefficient threshold $\tau>0$ and regularization parameters $\lambda_{\Omega}, \lambda>0$.
		\Ensure Support estimate $\hat{\mathcal{S}} \subseteq [d]$.
		\smallskip
		\Statex \textbf{At each local machine $m=1,\ldots,M$}
		\State Perform steps 1-5 as in Algorithm \ref{alg:count_votes}. 
		\setcounter{ALG@line}{5}		
		\State Set $\mathcal S^{m} = \left\{ \left(i, \mathrm{sign}(\hat{\xi}_i^m) \right) \,\Big|\, |\hat{\xi}_i^m|>\tau \right\}$ and send it to the fusion center.
		\smallskip
		\Statex \textbf{At the fusion center}
		\State For each $i\in[d]$, compute $S_i\in\re$ as $S_i = \sum_{m=1}^M \mathrm{sign}(\hat{\xi}_i^m) I \left\{ \left(i, \mathrm{sign}(\hat{\xi}_i^m)\right) \in \mathcal{S}^m \right\}$.
		\State \Return $\hat{\mathcal{S}} = \big\{ i \in [d] \,\big|\, |S_{i}|>V_T \big\}$.
	\end{algorithmic}
\end{algorithm}

{
\begin{remark}
	In practice, the lasso regularization parameter $\lambda$ is often set via 
	cross-validation \cite[Sec. 2.5.1]{buhlmann2011statistics}.
For simplicity, Algorithm~\ref{alg:two_round_scheme} is presented assuming 
all machines use the same 
regularization parameter $\lambda$. 
As we illustrate in the simulation section, our scheme performs well also if each machine sets its $\lambda$ by cross-validation. 
This is in accordance with our theoretical results
in Section \ref{sec:theoretical_results}, which do not 
require all machines to use the same $\lambda$, but only that
$\lambda$ in each machine be in a certain range. 
\label{remark:lambdas_cross_validated_lasso}
\end{remark}
}

\subsection{Variants of Algorithm \ref{alg:count_votes} }
\label{subsec:variations_supp_estim}

Several modifications of Algorithm \ref{alg:count_votes} may offer improved performance. 
One variant is a top $L$ algorithm where each machine sends to the center the indices of the $L$ largest entries of $|\hat\xi^m|$, for some  $K\leq L \ll d$. 
A similar approach was proposed in \citet{amiraz2022distributed} for the simpler problem of sparse normal means estimation.  
One advantage of this variant is that its communication per machine is fixed and known a priori $O(L \ln d)$. This is in contrast to the above thresholding based scheme, whose communication per machine is random. 

A different variant, described in Algorithm~\ref{alg:count_signed_votes}, 
 is to use the signs of the largest coefficients.  
Each machine sends the message 
$\mathcal{S}^m = \big\{ \big(i, \mathrm{sign}(\hat{\xi}_i^m) \big) \,\big|\, |\hat{\xi}_i^m|>\tau \big\}$.
Next, the fusion center computes 
the sum of received signs for each index $i\in[d]$, 
\begin{align}
    S_i = \sum_{m=1}^M \mathrm{sign}(\hat{\xi}_i^m) I\left\{ \left(i, \mathrm{sign}(\hat{\xi}_i^m)\right) \in \mathcal{S}^m \right\}.
    \label{e:sum_signs}
\end{align}
The estimated support set are the indices $i$ with values $|S_i|> V_T$.
This algorithm uses a few more bits than a voting scheme. However, 
with a large number of machines, 
sums of signs can better distinguish between support and non-support coefficients. The reason is that at non-support indices $j\not\in\cal S$, 
the sum $S_j$ has approximately zero mean, unlike sums of votes $V_j$, 
whereas at support indices $i\in\mathcal{S}$ they have similar magnitudes $|S_i| \approx V_i$ since such indices are unlikely to be sent with the 
opposite sign of $\theta^*_i$. 
In the simulation section we illustrate the improved performance of a sign-based over a votes-based distributed scheme.

\section{Theoretical results}
\label{sec:theoretical_results}

In this section we present a theoretical analysis of our scheme, in terms of support recovery and mean squared error in estimating $\theta^*$. 
For our theoretical results, we assume that each machine $m\in[M]$ has $n$ i.i.d. samples $(\vX^m, Y^m) \in \re^{n\times (d+1)}$ from the model \eqref{e:linear_model}, where the rows of $\vX^m$ follow $N(0, \Sigma)$, and $\theta^*$ is $K$ sparse.
We further make the following assumptions. 
 

\begin{assump}
	The singular values of $\Sigma \in \re^{d \times d}$ are bounded from above and below, 
		\label{assump:gaussian_design_bounded_spectrum} 
\begin{align}
    0<C_{\min}<\sigma_{\min}(\Sigma)\leq \sigma_{\max}(\Sigma)<C_{\max} < \infty.
    \label{e:singular_values_Sigma}
\end{align}
	For simplicity of the analysis, we assume that $C_{\max} \geq 1$.
\end{assump}
\begin{assump}
\label{assump:covariance_bounded_diagonal_elements} 
The diagonal elements of $\Sigma$ satisfy $\Sigma_{i,i} \leq 1$, $\forall i\in[d]$.
\end{assump}
\begin{assump}
\label{assump:row-wise_sparsity} 
The matrix $\Omega = \Sigma^{-1}$ is row sparse, with at most $K_\Omega$ non-zero off-diagonal entries per row, 
\begin{align}
    \max_{i\in [d]} \left| \{j\in[d] \,\big|  \,\Omega_{ij}\neq 0, \, j\neq i\} \right| \leq K_{\Omega}. 
    \label{e:row-wise_sparsity}
\end{align}
\end{assump}
\begin{assump}
\label{assump:regularization_precision_estimator}
{
Each machine computes $\hat{\Omega}^m$ by Algorithm \ref{alg:decorr_matrix} with 
$\lambda_{\Omega}^{m} = \kappa_{\Omega}^m \sqrt{\frac{\ln d}{n} }$, where  $\kappa_{\Omega}^m$ satisfies
\begin{equation}
		\label{e:kappa_Omega}
		8\sqrt{C_{\max}} < \kappa_{\Omega}^m < \sqrt{ \frac{n}{(K_{\Omega} + 1) \ln d}}.
\end{equation}
}
\end{assump}
\begin{assump}
\label{assump:inverse_block_covariance}
For $C_0=(32 C_{\max}/C_{\min}) + 1$, 
and a constant $\rho > 0$, 
$$\max_{J\subseteq [d],\, |J|\leq C_0 K}\| \Sigma_{J,J}^{-1}\|_{\infty} \leq \rho.$$
\end{assump}
\begin{assump}
	\label{assump:sample_size_support_recovery} 
	The sample size in each machine is sufficiently large,
	\begin{align}
		n \geq c \cdot \max\{K, (K_{\Omega}+1)^2\} \ln d,
		\label{e:sample_size_support_recovery}
	\end{align}
	for some suitable constant $c = c(C_{\min}, C_{\max}, \rho, \kappa_{\Omega})$.
\end{assump}
\begin{assump}
	\label{assump:lasso_regularization} 
	{
	Each machine computes the lasso in Eq. \eqref{e:lasso_estimator} with regularization $\lambda^m = \kappa^m \sigma \sqrt{(\ln d)/n}$, where $\kappa^m \in [8, \kappa_{\max}]$ for some constant $\kappa_{\max}$.
	}
\end{assump}

Similar assumptions appeared in previous theoretical works on the lasso and debiased lasso. 
In particular, under assumptions \ref{assump:gaussian_design_bounded_spectrum}--\ref{assump:inverse_block_covariance}, \ref{assump:lasso_regularization} and a slightly weaker assumption than \ref{assump:sample_size_support_recovery}, namely that $n > c \cdot \max\{K, K_{\Omega} + 1\} \ln d$,
\citet[Theorem 3.13]{javanmard2018debiasing} proved that up to a small bias term,
$\hat{\theta}$ is Gaussian.
In more detail, $\sqrt{n}(\hat{\theta} - \theta^*) = Z + R$, where $Z | \vX \sim N\big( 0, \sigma^2 \hat{\Omega}\hat{\Sigma}\hat{\Omega}^{\top} \big)$ and with high probability, the bias term is bounded as $\| R \|_{\infty} \leq \sigma \delta_R$, where
\begin{align}
\delta_R = C \frac{\ln d}{\sqrt{n}} \left( \rho\sqrt{K} + \min\{ K, K_{\Omega}\} \right).
\label{e:bias_level}
\end{align}
Assumption~\ref{assump:sample_size_support_recovery} 
requires a larger number of samples, 
because our scheme uses the vector $\hat{\xi}$ 
of Eq. \eqref{e:standardized_debiased_lasso}, 
whose coordinates involve division by $(\hat{\Omega}^m \hat{\Sigma}^m (\hat{\Omega}^m)^{\top})_{i,i}^{1/2}$. Bounding this quantity requires $n \gtrsim K_\Omega^2 \ln d$.  

To derive guarantees for exact support recovery, we  make the following assumption on the signal-to-noise ratio (SNR).
	
\begin{assump}
\label{assump:thetamin}
The non-zero coefficients of $\theta^*$ are sufficiently large. Specifically, $|\theta_i^*| > \theta_{\min}$ for all $i\in\mathcal{S}$, where
for some parameter $r>0$, 
\begin{align}
\theta_{\min} =  \sigma\sqrt{\frac{4 r }{C_{\min}}\frac{\ln d}n}.
\label{e:thetamin}
\end{align}
\end{assump}

The scalar $r$ in Eq. \eqref{e:thetamin} can be viewed as an SNR level, as it controls how small $\theta_{\min}$ can be. 
A value $r \asymp \frac{1}{M}$, namely $\theta_{\min} = c \sigma\sqrt{\frac{\ln d}{nM}}$, is a sufficient condition for a centralized lasso estimator with access to all $Mn$ samples to recover the support of $\theta^*$ \citep{wainwright2009sharp}.
In a distributed setting, at such a low SNR, individual machines may not be able to recover the support. 
Yet,  with the same SNR scaling, possibly with a larger multiplicative constant $c$, 
a distributed scheme with $O(d)$ communication per machine is able to recover the support \citep[Corollary 14]{lee2017communication}. 

A key theoretical question in our communication-restricted setting, is whether the fusion center can still recover the support while having limited communication with the $M$ machines. 
Since for $r \gtrsim 1$ each machine may recover the support using its own $n$ samples, the relevant SNR regime is thus  $\frac{1}{M} \lesssim r \lesssim 1$.
The following theorem shows that if the SNR is high enough, then Algorithm~\ref{alg:count_votes} recovers the support $\mathcal S$ of $\theta^*$ with high probability.

\begin{theorem}
Let $\hat{\cal S}$ be the support set found by Algorithm \ref{alg:count_votes} with $V_T = \ln d$ and 
\begin{align}
	\tau = \sqrt{2 \ln d} \,\left(1 + \frac\epsilon2 \right),  
			\label{e:tau_large}
\end{align}
where $\epsilon \geq 2 \delta_R \sqrt{\frac{C_{\max}}{\ln d}}$ and $\delta_R$ is defined in Eq. \eqref{e:bias_level}. 
Assume that \ref{assump:gaussian_design_bounded_spectrum}--\ref{assump:thetamin}  hold, that dimension $d$ is sufficiently large and the number of machines $M$ satisfies
\begin{align}
120 \ln d \leq M \leq \frac{\ln d}{2e^3} \cdot d .
\label{e:M_support_recovery_large_tau}
\end{align}
If the SNR satisfies
\begin{align}
r \geq \left( 1 + \epsilon - \sqrt{\frac{1}{\ln d} \ln\left( \frac{M}{c(d,M) \cdot \ln d} \right)} \right)^2 ,
\label{e:SNR_support_recovery_large_tau}
\end{align}
with $c(d,M) = 40 \sqrt{\ln\left( \frac{M}{40 \ln d} \right)}$,  
then 
\begin{align}
\Pr(\hat{\mathcal{S}} = \mathcal{S}) \geq 1 - \frac{K+1}{d} .
\label{e:support_recovery_large_tau}
\end{align}
\label{theorem:support_recovery_large_tau}
\end{theorem}
Let us make some remarks regarding this theorem. 
The reason for the term $\epsilon/2$ in the threshold $\tau$ in \eqref{e:tau_large}
is 
to overcome the remaining bias in the debiased lasso estimator.  
The threshold $\tau$ 
guarantees that each machine sends only few indices to the fusion center. 
Hence, for a number of machines $M \propto \ln d$, 
according to \eqref{e:SNR_support_recovery_large_tau}, the SNR $r$ at which
Algorithm~\ref{alg:count_votes} is guaranteed to recover the support 
is high and close to $r=1$. More data distributed into a larger number of machines $M$ allows recovery at lower SNRs. 
For example, if $M\propto d^\beta \ln d$ with $\beta<1$, then as $d\to\infty$, support recovery is guaranteed for 
$r \geq (1-\sqrt{\beta} + \epsilon)^2$, whereas
if $M \propto d$, then the required SNR in Eq. \eqref{e:SNR_support_recovery_large_tau}  
scales as $r \gtrsim \left( \epsilon + \frac{\ln \ln d}{\ln d} \right)^2$. 
{
} In simple words, if the number of machines is large and each machine has enough samples, then with very little communication per machine it is still possible to detect the support at SNR values much lower than $r=1$.
Unfortunately, this SNR is still much higher than the centralized rate of $r=1/M$.

The next theorem presents a support recovery guarantee for a lower threshold $\tau$. Here the number of machines $M$ 
can be arbitrarily large,  and the threshold $V_T$ depends on $M$. This allows recovery at lower SNR values
as compared to Theorem \ref{theorem:support_recovery_large_tau}.

\begin{figure}[t]
	\centering
	\includegraphics[scale=.5]{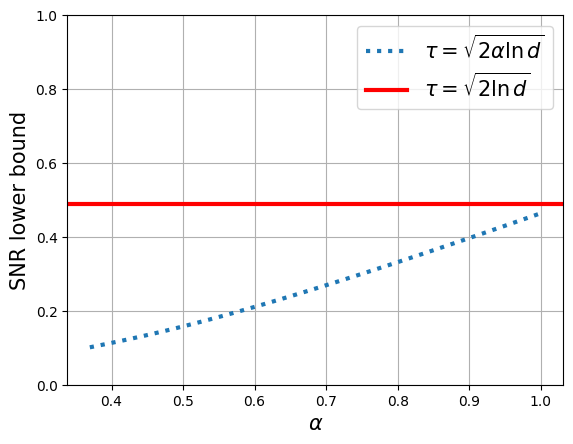}
	\caption{SNR lower bounds of Eq. \eqref{e:SNR_support_recovery_large_tau} (solid red) and \eqref{e:SNR_support_recovery_thresholds_not_fixed} (dotted) for $d=50000$, $M= 1133 \approx d^{0.65}$, $\epsilon=0$ and $V_T = \ln d$.}
	\label{fig:SNR_vs_threshold}
\end{figure}

\begin{theorem}
Assume that \ref{assump:gaussian_design_bounded_spectrum}--\ref{assump:thetamin} hold, that $d$ is sufficienly large and that the number of machines satisfies
\begin{align}
M = d^{\beta} \quad\text{for some}\quad
\beta > \frac{\ln(100 \ln d)}{\ln d} .
\label{e:M_support_recovery_thresholds_not_fixed}
\end{align}
For a fixed $\alpha\in\left(\frac{5}{\ln d},1\right]$, let $\hat{\cal S}$ be the support set found by 
Algorithm \ref{alg:count_votes} with  
\begin{align}
\tau = \sqrt{2 \ln d} \,\left(\sqrt{\alpha} + \frac\epsilon2 \right),
\label{e:tau_not_fixed}
\end{align}
where $\epsilon \geq 2 \delta_R \sqrt{\frac{C_{\max}}{\ln d}}$,
and with a voting threshold $V_T$ that satisfies
\begin{align}
2e d^{\beta - \alpha} e^{\frac{2\ln d}{V_T}} < V_T < {\frac{d^\beta}{100}}.
\label{e:alpha_VT_thresholds_not_fixed}
\end{align}
Then, for SNR sufficiently high, 
\begin{align}
r \geq \left( \sqrt{\alpha} + \epsilon - \sqrt{\frac{1}{\ln d} \ln\left( \frac{d^{\alpha + \beta}}{2(\ln d + V_T) d^{\alpha} + d^{\beta}} \cdot \frac{1}{c(\alpha, \beta, d, V_T)} \right)} \right)^2 ,
\label{e:SNR_support_recovery_thresholds_not_fixed}
\end{align}
where 
$c(\alpha, \beta, d, V_T) = 8\sqrt{\ln\left( \frac{d^{\alpha + \beta}}{16(\ln d + V_T) d^{\alpha} + 8d^{\beta}} \right)} $, 
\begin{align}
\Pr(\hat{\mathcal{S}} = \mathcal{S}) \geq 1 - \frac{K+1}{d} .
\label{e:support_recovery_thresholds_not_fixed}
\end{align}
\label{theorem:support_recovery_thresholds_not_fixed}
\end{theorem}
Figure \ref{fig:SNR_vs_threshold} illustrates the difference in the SNR lower bounds of the two theorems. 
For simplicity, we took $V_T = \ln d$ in both bounds, and set $\epsilon=0$ which corresponds to a large number of samples per machine so the bias term in the debiased lasso is negligible. 
As seen in the figure, with a smaller $\tau$ (lower value of $\alpha$), 
support recovery is possible at SNR values much lower than one. 

\subsection{Communication cost}

As mentioned in Section \ref{sec:proposed_method}, 
the communication of the first round of our scheme
depends on the threshold $\tau$. 
With a high threshold as in \eqref{e:tau_large}, 
the probability of each machine to send a non-support index is at most $2/d$. Hence if $\theta^*$ has sparsity $K$
the average communication per machine is 
$O(K \ln d)$.  In contrast, with a lower threshold
as in \eqref{e:tau_not_fixed}, the probability 
is at most $2/d^{\alpha}$. In this case,  
the average communication per machine is at most $O((K+d^{1-\alpha})\ln d )$.

Next, we consider the communication cost of the second round, and thus of the full two round scheme. 
Here, we make a mild assumption that 
$\max_i |\theta^*_i| = o(Mn)$. 
Under the assumptions 
made in the respective Theorems and this additional condition,  
the two round scheme that estimates $\theta^*$ (Algorithm \ref{alg:two_round_scheme}), has a communication
		cost comparable to that of the first round. 
		Indeed, with high probability, the exact support is recovered after the first round.
		Next, in the second round, each least-squares estimate $\hat{\beta}_i^m$ 
		can be accurately transmitted using $O(\log_2(nM))$ bits via Algorithm 1 of \citet[Lemma 2.3]{szabo2020adaptive} provided that $\mathbb{E}(\max\{1, \log_2 |\hat{\beta}_i^m|\}) = o\big(\log_2 (nM)\big)$.
		Hence, if $n$ and $M$ are at most polynomial in $d$, and the coefficients $\theta_i^*$ and thus
		also $\hat\beta_i^m$ are not extremely large, then  
		with high probability 
		the communication of the second round is $O\big( K\ln d \big)$ bits per machine.

\begin{remark}
	There is a tradeoff between the communication and the SNR at which support recovery is guaranteed. As $\alpha$ increases towards one, the communication cost decreases, but the SNR at which support recovery is guaranteed increases.  
	Furthermore, as $\alpha\nearrow 1$  
	the communication decreases exponentially fast towards $K\ln d$, which is the minimal cost to send the $K$ support indices. 
\end{remark}

We remark that deriving concentration bounds on the communication per machine is non-trivial. The reason is that
the coordinates of the debiased lasso estimate are correlated in a complicated manner, due to the inverse matrix $\Omega$ as well as the bias term. Hence, the total communication per machine in the first round is the sum of dependent Bernoulli random variables. As we show empirically in Section
\ref{sec:simulations}, the total communication per machine is tightly concentrated around its mean, implying that the dependence is weak. 
Finally, as also mentioned in Section \ref{sec:proposed_method}, from a practical perspective, a \texttt{top-L} variant, in which each machine sends $L$ coordinates in the first round may be preferable, as its first round communication is fixed and a-priori known.

\subsection{Accuracy of $\hat\theta$}

Beyond support recovery, 
another quantity of interest is the 
$\ell_2$ error $\|\hat\theta-\theta^*\|_2$ of $\hat\theta$. 
The next corollary shows that for a sufficiently high SNR, 
Algorithm~\ref{alg:two_round_scheme} achieves the same error rate as the oracle least squares estimator $\hat\theta^{\mbox{\tiny LS}}$ computed with the  samples
of all $M$ machines and with knowledge of the true support.

\begin{corollary}
Assume the conditions of Theorem \ref{theorem:support_recovery_large_tau} hold. Let $N=nM$ denote the total sample size of all $M$ machines. If $M = O\left( \frac{N K}{( \max\{K,\ln N\})^2} \right)$, then
the estimate $\hat{\theta}$ of Eq. \eqref{e:theta_hat}
satisfies that as $d,n\to\infty$, 
\begin{align}
\|\hat{\theta} - \theta^*\|_2 = O_{P}\left( \sqrt{\frac{K}{N}} \right).
	\label{eq:L2_error_oracle}
\end{align}
\label{corol:supprec_threshold_votes}
\end{corollary}
For the same distributed setting as ours, albeit without communication constraints,  
\cite{battey2018distributed} proposed a two-round estimator that attains the same rate as in Corollary \ref{corol:supprec_threshold_votes}. However, their
scheme incurs a communication of at least $O(d)$ bits,
as each machine sends $d$ values to the fusion center. 
In contrast, $\hat{\theta}$ in Algorithm \ref{alg:two_round_scheme} is computed with a much lower communication cost.

Finally, we note that 
	a lower bound on the magnitude of 
	the non-zero coefficients of $\theta^*$ 
	is a key requirement for our scheme to 
	satisfy \eqref{eq:L2_error_oracle}.
In Appendix \ref{sec:minimax_rate}, we prove
that without this requirement, our communication-constrained scheme does not achieve the minimax error rate. We conjecture that if the support coefficients can have arbitrarily small values, then the centralized minimax error rate is not achievable by any scheme whose communication is $O(d^{\gamma})$ for $\gamma <1$.

\subsection{Comparison to theoretical results of 
\citet{amiraz2022distributed}}

\cite{amiraz2022distributed} studied distributed estimation in the simpler sparse normal means problem. 
Their setting can be viewed as a specific instance of ours, with $n=d$ and design matrix $\vX^m=\vI_d$. In this case $Y^m$ is an unbiased estimator of $\theta^*$ and so there are no bias terms in their support recovery guarantees. 
We now show that if in our setting the number of samples $n$ is sufficiently large so that the bias $\epsilon$ is negligible, then our Theorems \ref{theorem:support_recovery_large_tau} and \ref{theorem:support_recovery_thresholds_not_fixed} 
have SNR requirements similar to those of \citet{amiraz2022distributed} for the sparse normal means problem.

Specifically, consider $\tau = \sqrt{2\ln d}$ and $M=d^\beta \ln d$ with $\beta < 1$. 
Then \citet[Theorem 2.C]{amiraz2022distributed} guarantees exact support recovery for sparse normal means at SNR $r\geq (1-\sqrt{\beta})^2$. In our case, for $d\gg 1$, Eq.~\eqref{e:SNR_support_recovery_large_tau} reads $r \gtrsim (1+\epsilon - \sqrt{\beta})^2$, where the term $\epsilon$ arises from the bias in the debiased lasso estimator. 
Hence, up to the bias $\epsilon$, the two results match each other. 
Next, \citet[Theorem 2.A]{amiraz2022distributed}
states that for $\tau = \sqrt{2 \alpha \ln d}$ and $M \propto \ln d$, support recovery 
is guaranteed for  $r\gtrsim \alpha$ as long as $\alpha \gtrsim \frac{1}{\ln d}$.  
As described in Section \ref{sec:comparison_to_Amiraz}, 
for similar values of $M$ and $\alpha$,  Eq.~\eqref{e:SNR_support_recovery_thresholds_not_fixed}
of Theorem~\ref{theorem:support_recovery_thresholds_not_fixed}
 yields $r \gtrsim \left( \epsilon + \frac{1}{\sqrt{\ln d}} \right)^2$, which matches their result as $\epsilon\to 0$.


\section{Simulations}
\label{sec:simulations}

We compare our approach to more communication intensive schemes via several simulations. 
We focus on the following methods, all based on debiased lasso: 
\begin{itemize}
    \item \texttt{thresh-thresh}: Algorithm \ref{alg:count_votes}, where the sparsity $K$ is unknown.
    \item \texttt{thresh-topK}: Similar to Algorithm \ref{alg:count_votes}, but the fusion knows the sparsity $K$ and sets $\hat{\mathcal S}$ to the $K$ indices with largest number of votes.
    \item \texttt{top5K-topK}: Both machines and fusion center know $K$. Each machine send the indices of its top $5K$ standardized coefficients. The fusion center sets $\hat{\mathcal S}$ to the 
    $K$ indices with largest number of votes. 
    \item \texttt{AvgDebLasso}: Based on \cite{lee2017communication}, each machine sends its debiased lasso estimate $\hat{\theta}^m$ to the fusion center. The center computes $\hat{\theta}^{avg} = \frac{1}{M} \sum_{m=1}^M \hat{\theta}^m$ and sets $\hat{\mathcal S}$ as the  $K$ indices with largest values  $|\hat{\theta}^{avg}_i|$.
    \end{itemize}
In the above methods, 
each machine computes a debiased lasso estimator using its own data. For a fair comparison, we run all methods with the same lasso regularization parameter $\lambda = 8 \sigma \sqrt{(\ln d)/n}$. Also, all machines
compute their 
estimate $\hat{\Omega}$ of $\Sigma^{-1}$ with regularization $\lambda_{\Omega} = 8\sqrt{C_{\max} (\ln d)/n}$. 
Note that only \texttt{thresh-thresh} does not know the sparsity $K$ and its fusion center uses a voting threshold
$V_T = \ln d$. Both \texttt{thresh-thresh} and \texttt{thresh-topK}  were run with a threshold 
$\tau = \sqrt{2\ln d}$. 
As a benchmark, we also ran a centralized estimator denoted
\texttt{oracle}. It knows the
exact support $\mathcal{S}$ and computes  $\hat\theta^{\mbox{\tiny LS}}$,  the least squares
estimate of $\theta^*$ using the data of all $M$ machines.
Finally, to illustrate the advantages of distributed inference over estimation using only the data at a single machine, we ran a scheme denoted \texttt{single} which computes a debiased lasso on only $n$ samples, followed by selection of its top-$K$ largest coefficients.

We evaluate the accuracy of 
an estimated support set $\hat{\mathcal{S}}$ by the F-measure, 
\begin{align*}
    \text{F-measure} = \frac{2\cdot \mathrm{precision} \cdot \mathrm{recall}}{\mathrm{precision} + \mathrm{recall}},
\end{align*}
where $\mathrm{precision}=|\mathcal{S}\cap \hat{\mathcal{S}}|/|\hat{\mathcal{S}}|$ and $\mathrm{recall}=|\mathcal{S}\cap \hat{\mathcal{S}}|/K$.
If $\hat S=\emptyset$ then we set $\text{F-measure}=0$. 
 An F-measure equal to one indicates exact support recovery.

Given a support estimate $\hat{\mathcal{S}}$, the vector $\theta^*$ is estimated as described in 
Section~\ref{sec:proposed_method}. 
For all methods excluding \texttt{oracle}, \texttt{AvgDebLasso} and \texttt{single}, we perform a second round and compute the estimator $\hat{\theta}$ using OLS restricted to the indices in the estimated support. 
The error of an estimate $\hat\theta$ is measured by 
$\|\hat{\theta} - \theta^*\|_2$.

\begin{figure}[t]
	\centering
	\includegraphics[width=\linewidth]{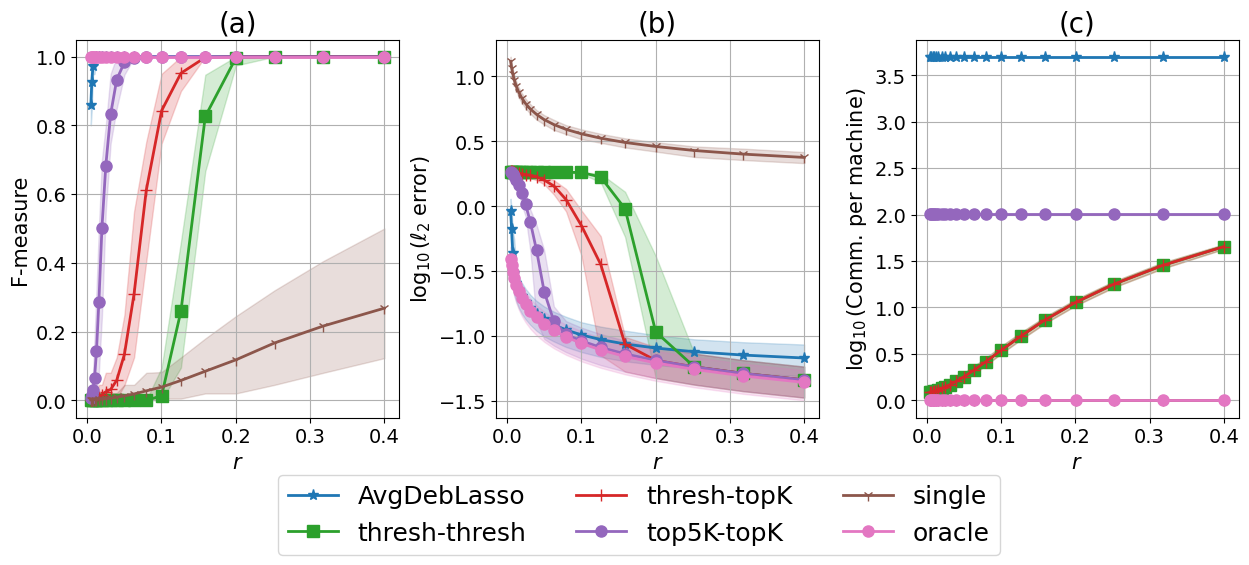}
	\caption{Simulation results for $n=250$, $d=5000$, $K=20$ and $M=100$. The shaded regions represent $90\%$
		confidence bands. 
	}
	\label{fig:varying_SNR_unknownPrec}
\end{figure}

We generated data as follows. 
The design matrix $\vX^m\in \re^{n\times d}$ in machine $m$ has $n$ rows i.i.d. $N(0,\Sigma)$, with $\Sigma_{i,j}=0.1^{|i-j|}$.
Next, we generated a $K$ sparse vector $\theta^*\in\re^d$, 
whose nonzero indices are sampled uniformly at random from $[d]$. 
Its nonzero coefficients are $\pm \theta_{\min}$ with random signs and magnitude $\theta_{\min} = \sqrt{(4 \ln d)/( C_{\min} n)}$, where $C_{\min}$ is the smallest singular value of $\Sigma$.
For a simulation with SNR parameter $r$, we set the noise level as $\sigma = 1/\sqrt{r}$. 
Finally, we generated the response $Y^m\in\re^n$ according to the linear model in Eq. (1).

Our first simulation compared the performance of various schemes as a function of the SNR, under the following setting: dimension $d=5000$, sparsity $K=20$, sample size in each machine $n=250$, and number of machines $M=100$. The SNR values considered lie in the interval $[\frac{1}{2M}, 0.4]$.
Figure \ref{fig:varying_SNR_unknownPrec}a displays the F-measure of each method, averaged over $100$ realizations. 
First of all, the
scheme \texttt{single} that uses data of only one machine is unable to achieve exact support recovery in this 
SNR range, as expected from theory.
At the lowest SNR values, \texttt{AvgDebLasso} achieved the best performance in terms of support recovery.
However, for stronger signals with $r>0.2$, all variants of our proposed approach achieved an F-measure of one, 
including \texttt{thresh-thresh}, which does not know the sparsity.
Additionally, at sufficiently high SNR, our methods estimate the support as accurately as \texttt{AvgDebLasso}, 
but with 2-3 orders of magnitude less communication, 
as shown in Figure~\ref{fig:varying_SNR_unknownPrec}c.
This plot shows the logarithm of the average number of indices sent per machine in the first round. Empirically, we found the variability to be extremely low. For example, at SNR $r=0.25$, the average number of indices sent was about 16 with a standard deviation smaller than one. 

Figure \ref{fig:varying_SNR_unknownPrec}b shows the errors $\|\hat \theta - \theta^*\|_2$ averaged over 100 realizations. At low SNR values close to {$1/M$}, \texttt{AvgDebLasso} has the error closest to the oracle.
However, for $r$ sufficiently higher than $1/M$, all variants of the proposed method yield estimates more accurate than \texttt{AvgDebLasso}. 
Indeed, at these SNR levels, our methods exactly recover the support, and the second round reduces to a distributed ordinary least squares restricted to the correct support set $\cal S$. In accordance with Corollary~\ref{corol:supprec_threshold_votes}, Algorithm~\ref{alg:two_round_scheme} has the same error rate as the oracle.

\begin{figure}[t]
	\centering
	\includegraphics[width=\linewidth]{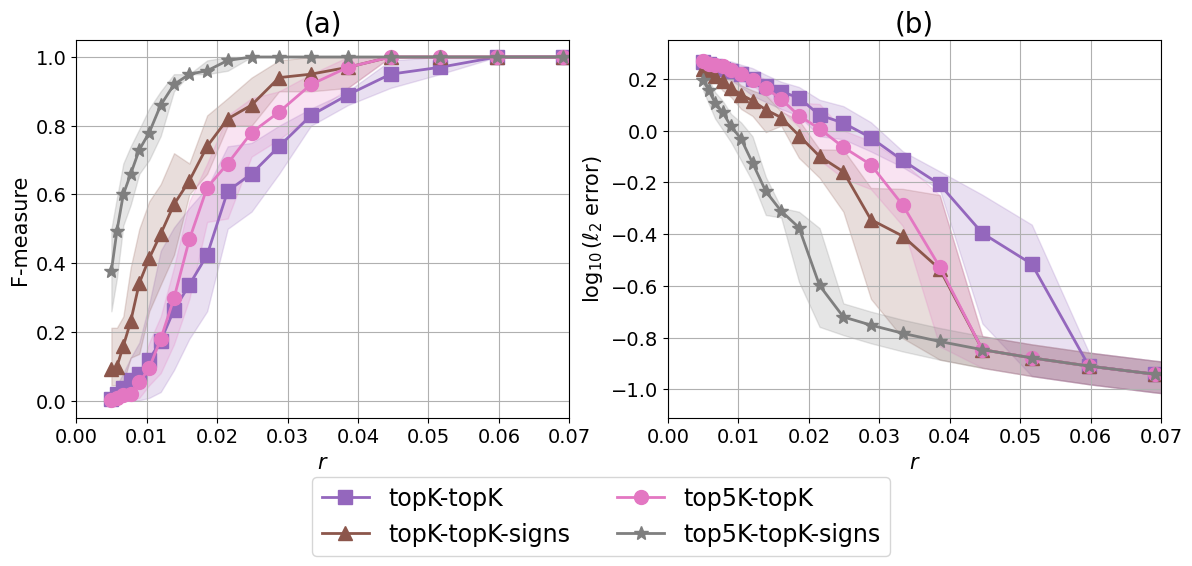}
	\caption{Simulation results for \texttt{topL-topK} and its variant \texttt{topL-topK-signs} where the first round uses sums of signs as described in Algorithm~\ref{alg:count_signed_votes}. 
}
	\label{fig:varying_SNR_knownPrec_signs}
\end{figure}

\subsection{Advantage of sending signs}

The following simulation compares sums of signs (Algorithm~\ref{alg:count_signed_votes})
to sums of votes (Algorithm~\ref{alg:count_votes}), in terms of support recovery and parameter estimation. 
As in the previous simulation, 
$n=250$, $d=5000$, $M=100$ and $K=20$.
Figure \ref{fig:varying_SNR_knownPrec_signs} shows that sums of signs is more accurate than sums of votes for low SNR values in the range $r\in[1/M, 0.1]$.
Figure \ref{fig:varying_SNR_knownPrec_signs} also shows that using $L=5K$ instead of $L=K$ significantly improves the accuracy of the support estimator, at the expense of increasing the communication. 
Hence, a small increase in the communication can improve considerably the accuracy of support and parameter estimators.

\begin{figure}[t]
	\centering
	\includegraphics[width=\linewidth]{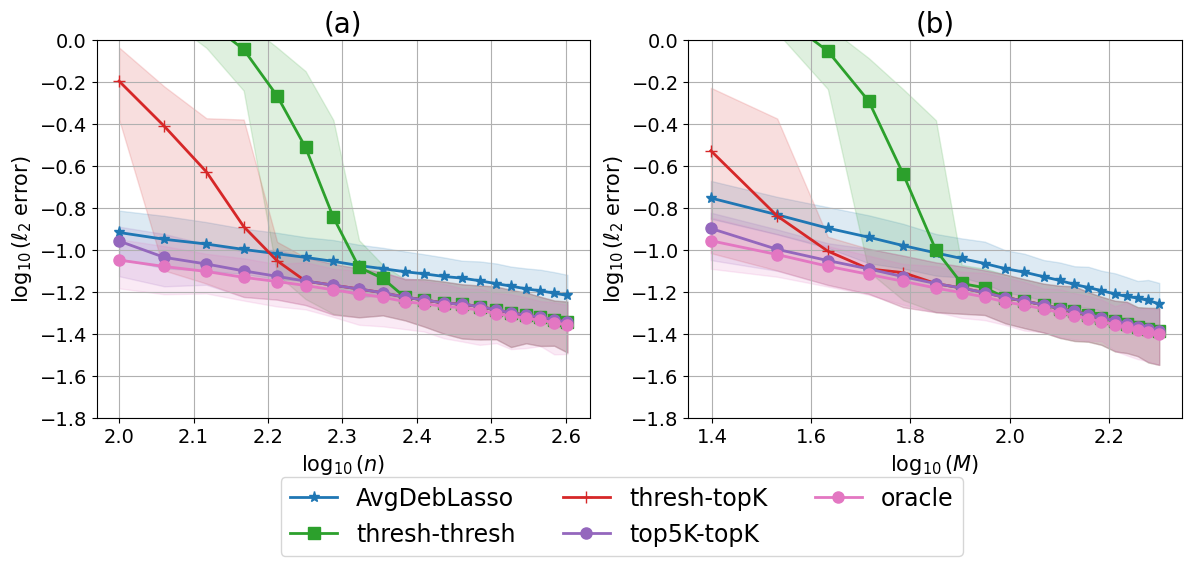}
	\caption{Estimation error (on a log-log scale) as a function of sample size $n$ or number of machines $M$. In both cases, $d=5000$, $K=20$ and $\theta_{\min} = 0.4081$.}
	\label{fig:varying_n_M}
\end{figure}

\subsection{Error decay versus sample size or number of machines}

Here we studied the dependence of the error 
$\| \hat\theta - \theta^*\|_2$
as a function of sample size $n$ or number of machines $M$. 
To this end, we considered a setting where $n$ varies in the interval $[100,400]$ with $M=100$ fixed, and another setting where $M$ varies in the interval $[25, 200]$ with $n=250$ samples per machine.
We fixed $d=5000$, $K=20$ and $\theta_{\min} = 0.4081$ (corresponding to $r=0.25$ at $n=250$). 
To save on computational time, in these simulations we assumed the matrix $\Omega$ was a-priori known. 
Figure~\ref{fig:varying_n_M} shows the resulting $\ell_2$ errors averaged over 100 simulations.
Note that for sufficiently large $n$ or $M$, the estimation errors of the proposed methods are close to the oracle.
Indeed, larger $n$ increases the effective SNR whereas increasing $M$ allows the methods to detect weaker signals, eventually leading to an estimation error close to that of the oracle.
The plots in Figure~\ref{fig:varying_n_M} show a linear dependence on a log-log scale with a slope of approximately $-1/2$, implying that the resulting errors decay as $1/\sqrt{n}$ and $1/\sqrt{M}$, respectively. 
This is in agreement with the theoretical result in Corollary~\ref{corol:supprec_threshold_votes}.


\subsection{Cross Validation}

\begin{figure}[t]
	\centering
	\includegraphics[width=\linewidth]{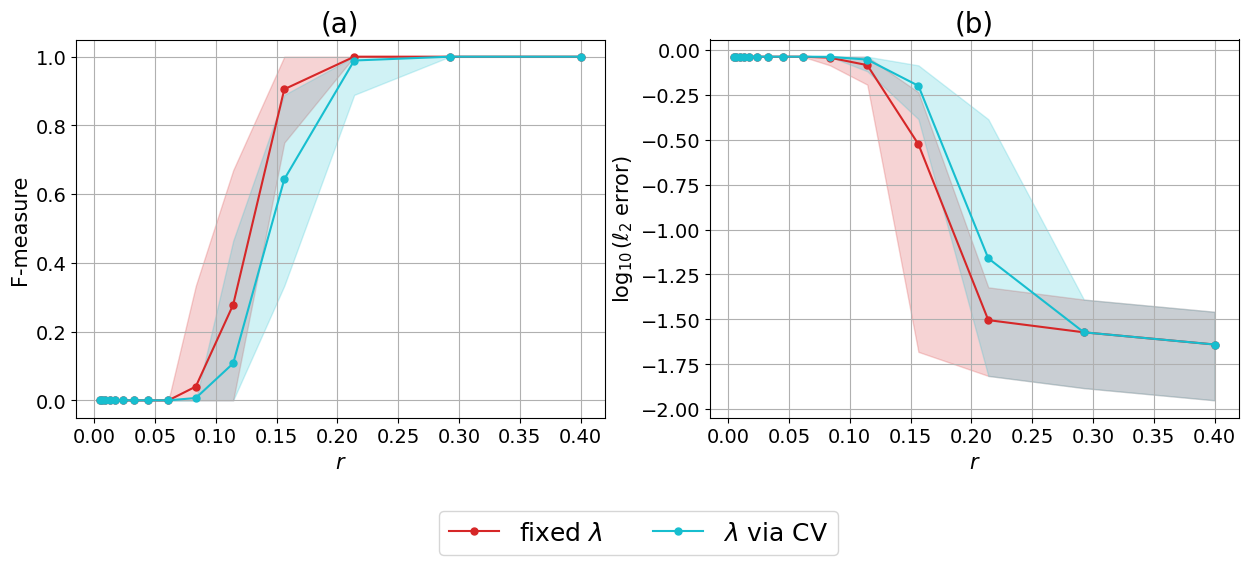}
	\caption{Accuracy of support recovery (left) and error $\|\hat \theta-\theta^*\|_2$ on  a log scale (right),  as function of SNR, averaged over 50 realizations.
		Red dots are results with a fixed identical $\lambda$ in all machines. 
		Cyan dots are results with lasso parameter $\lambda$ estimated separately at each machine by 10-fold cross validation. }
	\label{f:fmeas_risk_vary_SNR_cv}
\end{figure}

In this section we illustrate that our scheme continues
to perform well if individual machines each choose their
 regularization $\lambda$ via cross-validation. 
 In this simulation $d=1000$, $n=200$, $M=100$ and $K=5$. Our scheme was applied with $\tau = \sqrt{2\ln d}$ and $V_T = \ln d$. 
All machines used a fixed 
$\lambda_\Omega = 8\sqrt{ C_{\max}(\ln d)/n}$ to estimate their precision matrices.
We considered two options for the lasso regularization $\lambda$: (i) fixed at $\lambda=8\sigma\sqrt{(\ln d)/n}$;
or (ii) each machine separately selected $\lambda$ via 10-fold cross-validation on its prediction error. 
Figure~\ref{f:fmeas_risk_vary_SNR_cv} shows that
both options yield similar performance, 
and thus illustrate 
that our scheme is not too sensitive to the choice of regularization parameter.

\section{Summary and Discussion}
    \label{sec:summary}

The development and statistical analysis of distributed schemes with limited communication are important contemporary problems.
Given its simplicity and ubiquity, distributed sparse linear regression has attracted significant attention. 
As detailed in \citet[Table 1]{ji2023distributed}, most distributed sparse regression schemes require a communication of at least $O(d)$ bits per machine.
In contrast, 
in this work, we proved theoretically and showed via simulations that, under suitable conditions, accurate distributed inference for sparse linear regression is possible
with a much lower communication per machine, that is sublinear in $d$.

Over the past years, several authors studied distributed statistical inference under communication constraints. Specifically, for sparse linear regression, 
\cite{braverman2016communication} proved that without a lower bound on the SNR, to obtain a risk comparable to that of the minimax lower bound, a communication of at least $\Omega(M \min(n,d)/\log d)$ bits is required. 
\citet{acharya2019distributed} proved that, under certain conditions, rate optimal estimates of a sparse linear regression model cannot be computed using total communication sublinear in the dimension. 
However, as mentioned in their appendix B.3, a precise characterization of the ability to recover 
the support 
$\mathcal{S}$ with sublinear communication in $d$ and its dependency on other parameters (SNR, $M$, etc.) is still an open problem. 
In our theoretical results, we presented explicit expressions for the minimal SNR at which our scheme is guaranteed to achieve exact recovery with high probability and with sublinear communication in $d$. 
Our results highlight potential tradeoffs between SNR, communication and number of machines.

It may be possible to extend and improve
our theoretical analysis
with more refined techniques.
For example, since the $d$ coordinates of a debiased lasso estimator are  correlated, sharp concentration bounds for dependent variables, like those of \cite{lopes2022sharp}, could improve the  analysis and extend it to other schemes such as top-$L$.
Another extension is to relax the assumption that
both the noise and covariates follow Gaussian distributions. \citet{lee2017communication} and \citet{battey2018distributed}, for example, considered sub-Gaussian distributions for these terms, and other more general models.  
Another possible extension 
	is a theoretical analysis of the advantages of the sending signs scheme in Algorithm~\ref{alg:count_signed_votes}. 

Finally, our low-communication schemes could also be applied to other problems, such as sparse $M$-estimators, sparse covariance estimation and distributed estimation of jointly sparse signals. We leave these for future research.

\begin{appendix}

\section{Auxiliary Lemmas}
\label{sec:auxiliary_results}


\subsection{Bounds on the error of the lasso} 

We review several results on the lasso, used in our analysis.
A key requirement on the design matrix $\vX$ is the  
{\em restricted eigenvalue condition}, which bounds 
$\|\vX v\|$
over
approximately sparse vectors $v$.  
As in \citet{javanmard2018debiasing}, for an integer $K$ and $L>0$, consider 
the following subset of 
$\re^d$:
\begin{align}
\mathcal{C}(K,L,d) = \left\{ v\in\re^{d} \,\big|\, \exists \mathcal{S} \subseteq [d], |\mathcal{S}|=K, 
\|v_{\mathcal{S}^c}\|_1 \leq L\|v_{\mathcal{S}}\|_1 \right\}.
\label{e:set_cone_constraints}
\end{align}
The next result \citep[Lemma 3.2]{javanmard2018debiasing} 
presents conditions under which $\vX$ satisfies the
restricted eigenvalue condition on $\mathcal{C}(K,L,d)$. 

\begin{lemma}
Let $\vX\in\re^{n\times d}$ have i.i.d. rows from $N(0,\Sigma)$ where $\Sigma$ satisfies assumption \ref{assump:gaussian_design_bounded_spectrum}. For fixed $0<\delta<1$, $0<K<d$ and $L>0$, define
\begin{align}
	\!\!\!\!
	\mathcal{B}_{\delta}(n,K,L) \!=\! \left\{ \vX \in \re^{n\times d} \,\Bigg|\,
	\begin{aligned}
		&(1-\delta)\sqrt{C_{\min}} \leq \frac{\|\vX v\|_2}{\sqrt{n}\|v\|_2} \leq (1+\delta)\sqrt{C_{\max}}, \\
		&\forall v\in \mathcal{C}(K,L,d) \text{ s.t. } v\neq 0 
	\end{aligned}
	\right\}.
	\label{e:set_Bdelta}
\end{align}
Then there exists $ c_1 = c_1(L) > 0$, such that for $n > c_1 K \ln(d/K)$, 
\begin{align*}
\Pr \left( \mathcal{B}_{\delta}(n,K,L) \right) \geq 1 - 2e^{-\delta^2 n}.
\end{align*}
\label{lemma:RudelsonZhou2013_teo16}
\end{lemma}

A common approach to bound the error of the lasso, is to exclude a low-probability event where the noise is highly correlated to a non-support column of $\vX$. The next lemma bounds the probability of this event. 
It is similar to \citet[Lemma 6.2]{buhlmann2011statistics}, which assumed $\hat\Sigma_{i,i}=1$ $\forall i$. 
\begin{lemma}
	\label{lemma:prob_tildeB}
	Let $\vX\in\re^{n\times d}$ have i.i.d. rows from $N(0,\Sigma)$ and assume $W\sim N(0,\sigma^2 \vI_{n})$ is independent of $\vX$. Consider the event
	\begin{align}
		\tilde{\mathcal{B}} = \tilde{\mathcal{B}}(n,d,\sigma) = 
		\left\{ 
		\| 
		\vX^{\top}W \|_{\infty} \leq 4\sigma\sqrt{\tfrac{\ln d}{n}} \right\}.
		\label{e:set_Btilde}
	\end{align}
	Then under assumption  \ref{assump:covariance_bounded_diagonal_elements},   
	$
	\Pr\big( \tilde{\mathcal{B}} \,\big) \geq 1 - \frac{2}{d^3} - d \cdot e^{-n/8}. 
	$
\end{lemma}


The next lemma presents error bounds for the lasso estimator, assuming that the design matrix $\vX$ satisfies the restricted eigenvalue condition, and the noise is not too correlated to any column of $\vX$, namely under $\mathcal{B}_{\delta}(n,K,3) \cap \tilde{\mathcal{B}}(n,d,\sigma)$.

\begin{lemma}
Let	$Y=\vX \theta^* + \sigma W$, 
where $W\sim N(0,\vI_{n})$ is independent of $\vX$, and $\theta^*$ is $K$-sparse. 
Further assume the rows of $\vX\in \re^{n\times d}$ are i.i.d.  $N(0,\Sigma)$, where $\Sigma$
satisfies assumption  \ref{assump:gaussian_design_bounded_spectrum}. 
Let $\lambda \geq 8\sigma \sqrt{\frac{\ln d}{n}}$ and $0<\delta<1$. Under the event 
$\mathcal{B}_{\delta}(n,K,3) \cap \tilde{\mathcal{B}}(n,d,\sigma)$,
the lasso estimator $\tilde{\theta} = \tilde{\theta}(\vX, Y, \lambda)$ 
of Eq. \eqref{e:lasso_estimator}
satisfies
\begin{align}
\| \tilde{\theta} - \theta^* \|_1 \leq \frac{8 K }{(1-\delta)^2 C_{\min}}  \lambda
\label{e:lasso_ell1Error}
\end{align}
and
\begin{align}
\frac{1}{n} \| \vX (\tilde{\theta} - \theta^* ) \|_2^2 \leq \frac{16 K}{(1-\delta)^2 C_{\min}}  \lambda^2 .
\label{e:lasso_ell2predictionError}
\end{align}
\label{lemma:lasso_ell1Error_ell2predictionError}
\end{lemma}
For completeness, the proof of Lemma \ref{lemma:lasso_ell1Error_ell2predictionError} appears in appendix \ref{sec:proof_auxiliary_lemmas}. It follows \citet[Section 6.2.2]{buhlmann2011statistics}, 
albeit with different constants as their objective $\frac{1}{n}\|Y - \vX\theta\|_2^2 + \lambda\|\theta\|_1$, does not have a factor of $1/2$ as in \eqref{e:lasso_estimator}.


The next lemma bounds the $\ell_2$ error
 \citep[Theorem 7.13(a)]{wainwright2019high}.

\begin{lemma}
Let the rows of $(Y,\vX) \in \re^{n\times (d+1)}$ be i.i.d. from the model~\eqref{e:linear_model}, 
where $\theta^*$ is $K$ sparse.
Assume that $\vX \in \re^{n\times d}$ satisfies the restricted eigenvalue condition over $\mathcal{C}(K,3,d)$
with constant $\phi>0$, i.e., $\frac{1}{n}\| \vX v \|_2^2 \geq \phi \|v\|_2^2$ for all $v \in \mathcal{C}(K,3,d)$.
Then, for any solution $\tilde{\theta}$ of 
\eqref{e:lasso_estimator} with regularization $\lambda \geq 2\|\frac{1}{n}\vX^{\top}W\|_{\infty}$,
\begin{align*}
\| \tilde{\theta} - \theta^* \|_2 \leq \frac{3 \sqrt{K}}{\phi} \lambda .
\end{align*}
\end{lemma}
\begin{corollary}
Let the rows of $(Y,\vX) \in \re^{n\times (d+1)}$ be i.i.d. from the model~\eqref{e:linear_model}, where $\theta^*$ is $K$ sparse and $\Sigma$ satisfies \ref{assump:gaussian_design_bounded_spectrum}.
Under $\mathcal{B}_{\delta}(n,K,3) \cap \tilde{\mathcal{B}}(n,d,\sigma)$, it follows that  $\phi \geq (1-\delta)^2 C_{\min}$ and  $\|\frac{1}{n}\vX^{\top}W\|_{\infty} \leq 4\sigma \sqrt{\frac{\ln d}{n}}$. Hence, for any $\lambda \geq 8\sigma \sqrt{\frac{\ln d}{n}}$, 
\begin{align}
\| \tilde{\theta} - \theta^* \|_2 \leq \frac{ 3\sqrt{K} }{ (1-\delta)^2 C_{\min} } \lambda .
\label{e:lasso_ell2Error}
\end{align}
\label{corollary:lasso_ell2Error}
\end{corollary}


\subsection{Results for the precision matrix estimator}

Recall that the estimator $\hat{\Omega}$ in Algorithm \ref{alg:decorr_matrix} is computed row by row.
Under the assumption that all rows of $\vX$ are i.i.d. $N(0,\Sigma)$, each column $X_j$ of $\vX$ can be written as a linear combination of its other $d-1$ columns $\vX_{-j} \in \re^{n\times (d-1)}$, with an additive Gaussian residual term \citep[Eq. (43)]{zhang2014confidence}. Namely,
\begin{align}
X_j = \vX_{-j}\gamma_j + \varepsilon_j,
\quad \varepsilon_j \sim N\left(0, \frac{1}{\Omega_{j,j}}\vI_{n}\right),
\label{e:model_rowwise_lasso}
\end{align}
where $\varepsilon_j$ is independent of $\vX_{-j}$, and the vector $\gamma_j\in\re^{d-1}$ is given by
\begin{align}
\gamma_j = \left( -\frac{\Omega_{j,1}}{\Omega_{j,j}}, \ldots, -\frac{\Omega_{j,j-1}}{\Omega_{j,j}}  \,,\, -\frac{\Omega_{j,j+1}}{\Omega_{j,j}}, \ldots, -\frac{\Omega_{j,d}}{\Omega_{j,j}} \right).
\label{e:gamma_j}
\end{align}
Assumption \ref{assump:row-wise_sparsity} implies that $\gamma_j$ is $K_\Omega$ sparse. Algorithm \ref{alg:decorr_matrix} estimates it by
\begin{align*}
\tilde{\gamma}_{j} = \arg\min_{\gamma\in\re^{d-1}} \left\{ \frac{1}{2n}\|X_j - \vX_{-j}\gamma\|_2^2 
+ \lambda_\Omega
\|\gamma\|_1  \right\},
\end{align*}
whereas the quantity $1/\Omega_{j,j}$ is estimated by
\begin{align}
\tilde{\tau}^2_j = \frac{1}{n}\|X_j - \vX_{-j}\tilde{\gamma}_{j}\|_2^2 
+ \lambda_\Omega
\|\tilde{\gamma}_{j}\|_1.
\label{e:tilde_tau}
\end{align}
Let  
$
\tilde{\vC}_j = (-\tilde{\gamma}_{j,1},\ldots,-\tilde{\gamma}_{j,j-1} \,,\, 1 \,,\, -\tilde{\gamma}_{j,j+1},\ldots, -\tilde{\gamma}_{j,d}).
$
Then, the $j$-th row of $\hat{\Omega}$ is estimated by
\begin{align}
\hat{\Omega}_j= \frac{1}{\tilde{\tau}^2_j} \tilde{\vC}_j 
= \frac{1}{\tilde{\tau}^2_j} (-\tilde{\gamma}_{j,1},\ldots,-\tilde{\gamma}_{j,j-1},1,-\tilde{\gamma}_{j,j+1},\ldots, -\tilde{\gamma}_{j,d}).
\label{e:lasso_row_estimator}
\end{align}
To derive error bounds for $\hat\Omega_j$, we define the following event for the matrix $\vX_{-j}$,
\begin{align}
	\mathcal{B}_j = \mathcal{B}_{\delta j}(n, K_{\Omega}, 3) \cap \tilde{\mathcal{B}}_j(n, d-1, \sqrt{C_{\max}}),
	\label{e:set_Btilde_delta}
\end{align}
where $\mathcal{B}_{\delta,j}$ is the analogue of $\mathcal{B}_{\delta}(\cdot)$ in Eq. \eqref{e:set_Bdelta}, but with the matrix $\vX_{-j}$
instead of $\vX$. Similarly, $\tilde{\mathcal B}_j$ is 
defined as follows, 
with $\gamma_j$ given in Eq. \eqref{e:gamma_j},  
\begin{align*}
	\tilde{\mathcal{B}}_j(n, d-1, R) = 
		\left\{
		\vX \in \re^{n\times d} \,\Bigg|\, 
		\| 
		\vX_{-j}^{\top}(X_j - \vX_{-j}\gamma_j) \|_{\infty} \leq 4R\sqrt{\frac{\ln(d-1)}{n}} \right\}.	
\end{align*}
The next lemma shows that under
$\mathcal{B}_j$, $\tilde{\tau}_j^2$ is close to $\frac{1}{\Omega_{j,j}}$, with high probability.
\begin{lemma}
Let $\vX$ have i.i.d. rows from $N(0, \Sigma)$. Assume \ref{assump:gaussian_design_bounded_spectrum}, \ref{assump:row-wise_sparsity} and \ref{assump:regularization_precision_estimator} and $n\geq 24{\ln d}$.
Then, for any $\delta\in(0,1)$ there exists $c = c(\delta, C_{\max}, C_{\min})$ such that 
\begin{align*}
\Pr \left( \left| \tilde{\tau}_j^2 - \frac{1}{\Omega_{j,j}} \right| \leq c \sqrt{K_{\Omega} + 1} \cdot \lambda_{\Omega} \,\Bigg|\, \mathcal{B}_j \right) \geq 1 - \frac{2}{d^3}.
\end{align*}
\label{lemma:concentration_tilde_tau_j}
\end{lemma}

\begin{remark}
The proof of Lemma \ref{lemma:concentration_tilde_tau_j} relies on Lemma \ref{lemma:lasso_ell1Error_ell2predictionError} on the accuracy of the lasso estimate $\tilde{\gamma}_j$. 
To invoke it we assume \ref{assump:gaussian_design_bounded_spectrum}, \ref{assump:row-wise_sparsity} and \ref{assump:regularization_precision_estimator} all hold. Specifically, by Eq. \eqref{e:model_rowwise_lasso},  the noise level in row $j\in[d]$ is $1/\Omega_{j,j}$. Hence, the condition $\kappa_\Omega > 8 \sqrt{C_{\max}}$ in \ref{assump:regularization_precision_estimator} implies that $\lambda_\Omega$ is sufficiently large to apply Lemma \ref{lemma:lasso_ell1Error_ell2predictionError}.
\end{remark}

Lemma \ref{lemma:concentration_tilde_tau_j} leads to the following
corollary: 
\begin{corollary}
Under the conditions of Lemma \ref{lemma:concentration_tilde_tau_j}, 
and for sample size  
$
n \geq \frac{4 c^2 \kappa_\Omega^2}{C_{\min}^2} (K_\Omega + 1) \ln d, 
$
where $c=c(\delta,C_{\min},C_{\max})$ is the constant in Lemma
\ref{lemma:concentration_tilde_tau_j},
\begin{equation}
\Pr\left(\frac{1}{\tilde{\tau}^2_j} \leq \frac{2}{C_{\min}} \,\Big|\,
\mathcal{B}_j \right)
 \geq 1 - \frac{2}{d^3}.
 	\label{eq:lower_bound_tilde_tau}
\end{equation}
\label{corollary:bound_inverse_tilde_tau_j}
\end{corollary}
This corollary is similar to \citet[Lemma C.1]{javanmard2018debiasing}.
{However, to the best of our understanding, their lemma
	is incorrect as stated, since it claims that for a sufficiently large sample size 
	$1/\tilde{\tau}_j^2 \leq 2/C_{\min}$ holds deterministically, rather than with high probability
	as in \eqref{eq:lower_bound_tilde_tau}.}

The next lemma states that under suitable conditions,
$\hat\Omega_j$ 
is close to $\Omega_j$.

\begin{lemma}
Let $\vX \in \re^{n\times d}$ have i.i.d. rows from $N(0,\Sigma)$.
Assume \ref{assump:gaussian_design_bounded_spectrum}, 
\ref{assump:row-wise_sparsity} and \ref{assump:regularization_precision_estimator} hold.
For $\delta \!\in\! (0,1)$ and $j\in[d]$, let $\mathcal{B}_j$ be the corresponding set defined in Eq. \eqref{e:set_Btilde_delta}. Assume $n\geq 24{\ln d}$
as well as 
$n\geq \frac{4 c^2 \kappa_\Omega^2}{C_{\min}^2} (K_\Omega + 1) \ln d$ as in Corollary~\ref{corollary:bound_inverse_tilde_tau_j}. 
Then, there exists $c_1=c_1(\delta, C_{\min}, C_{\max})$ and $c_2=c_2(\delta, C_{\min}, C_{\max})$ such that 
\begin{align}
\Pr\left(  \| \hat\Omega_j - \Omega_j \|_1 \leq 
 c_1 (K_\Omega+1)\lambda_\Omega 
\,\Bigg|\, \mathcal{B}_j \right) \geq 1 - \frac{2}{d^3},
\label{e:bound_ell1_error_Omegahat_j}
\end{align}
and
\begin{align}
\Pr\left( \| \hat\Omega_j - \Omega_j \|_2 \leq 
c_2 \sqrt{K_\Omega + 1} \cdot \lambda_\Omega
\,\Bigg|\, \mathcal{B}_j \right) \geq 1 - \frac{2}{d^3}.
\label{e:bound_ell2_error_Omegahat_j}
\end{align}
\label{lemma:bound_ell1_ell2_error_Omegahat_j}
\end{lemma}


Recall the decomposition of the debiased lasso, $\sqrt{n}(\hat \theta - \theta^*) = Z+R$, where $Z|X\sim N(0,\sigma^2\hat\Omega\hat\Sigma\hat\Omega^\top)$. 
The next lemma gives a non-asymptotic bound on the variance of the entries of $Z|X$. 
It is a refinement of Lemma 5.4 of \cite{van2014asymptotically}, which is asymptotic in nature.
\begin{lemma}
Let $j\in[d]$. 
	Under Assumptions \ref{assump:covariance_bounded_diagonal_elements},
\ref{assump:sample_size_support_recovery} and the conditions of Lemma \ref{lemma:bound_ell1_ell2_error_Omegahat_j},
for a suitable $c=c(C_{\min}, C_{\max})$, 
and with probability at least 
$1- 13/d^3 - 2(d+2)e^{-n/8}$, 
\begin{equation}
	 | (\hat{\Omega}\hat{\Sigma}\hat{\Omega}^{\top})_{j,j} - \Omega_{j,j} |  \leq c  \cdot (K_\Omega+1) \lambda_\Omega.
	\label{e:bound_variance_debiased_lasso}
	\end{equation}
\label{lemma:bound_variance_debiased_lasso}
\end{lemma}

\begin{corollary}
	Assume that Eq. (\ref{e:bound_variance_debiased_lasso}) holds. 
If the sample size satisfies $n > c \kappa_{\Omega}^2 (K_{\Omega} + 1)^2 \ln d$ for some suitable constant $c>0$, then
\begin{align}
\frac{1}{2 C_{\max}} \leq \left( \hat{\Omega}\hat{\Sigma}\hat{\Omega}^{\top} \right)_{j,j} \leq \frac{2}{C_{\min}}.
	\label{eq:bound_Omega_Sigma_Omega}
\end{align}
\label{corollary:bound_variance_debiased_lasso}
\end{corollary}


\subsection{Properties of the debiased lasso estimator}

The following theorem \citep[Theorem 3.13]{javanmard2018debiasing} is a key component in 
our proofs .
It states that the debiased lasso estimator follows a 
Gaussian distribution up to a bias term, and provides
a bound on this bias. 
\begin{theorem}
		Let the rows of $(Y,\vX) \in \re^{n\times (d+1)}$ be i.i.d. from the model~\eqref{e:linear_model}, with a $K$-sparse $\theta^*$.  
	Let $\tilde{\theta}$ be the lasso estimator computed using $\lambda = \kappa\sigma\sqrt{(\ln d)/n}$ for $\kappa\in [8, \kappa_{\max}]$, and let $\hat{\theta}$ be the debiased lasso estimator in Eq. \eqref{e:debiased_lasso} with $\hat{\Omega}$ computed by  Algorithm \ref{alg:decorr_matrix} with 
	$\lambda_{\Omega}=\kappa_{\Omega}\sqrt{(\ln d)/n}$ for some suitably large $\kappa_{\Omega}>0$.
			Assume that \ref{assump:gaussian_design_bounded_spectrum}, \ref{assump:covariance_bounded_diagonal_elements}, \ref{assump:row-wise_sparsity}, and \ref{assump:inverse_block_covariance} hold.
		Then there exist constants $c$, $c_*$, $C$ depending solely on $C_{\min}$, $C_{\max}$, $\kappa_{\max}$ and $\kappa_{\Omega}$ such that, for $n \geq c \max\{K, K_{\Omega}\}\ln d$, 
	\begin{align}
		\sqrt{n}(\hat{\theta} - \theta^*) = Z + R, \quad Z|\vX \sim N(0, \sigma^2 \hat{\Omega}\hat{\Sigma}\hat{\Omega}^{\top}), \label{e:deblasso_estimator_representation_Z_R}
	\end{align}
	where $\hat{\Sigma}=\vX^{\top}\vX/n$, 
	and with probability at least $1 - 2 d e^{-c_* n/K} - d e^{-c n} - 6 d^{-2}$, 
	\begin{equation}
		\|R\|_{\infty} \leq 
		\sigma \cdot C\frac{\ln d}{\sqrt{n}}\left(
		{\rho\sqrt{K}} + \min\{K, K_{\Omega}\}
		\right).
		\label{e:bound_bias_term_R}   
	\end{equation}    
\label{theorem:deblasso_asymp_dist}
\end{theorem}

The next lemma bounds the deviations of the normalized debiased lasso. 

\begin{lemma}
	For any $i\in[d]$, let
	\begin{align}
		\mathcal{C}_i = \left\{ \vX\in\re^{n\times d} \Big| \frac{1}{2 C_{\max}} \leq (\hat{\Omega}\hat{\Sigma}\hat{\Omega}^{\top})_{i,i} \leq \frac{2}{C_{\min}} \right\}.
		\label{e:set_Ci}
	\end{align}
	Further, define $p = 4d e^{-c_* n/K}+2de^{-cn} + 12/d^2$, where $c,c_*$ are the constants in Theorem \ref{theorem:deblasso_asymp_dist}. 
	Then, under the assumptions of Theorem \ref{theorem:deblasso_asymp_dist}, 
	with $n$ and $d$ sufficiently large
	and $\delta_R$  defined in Eq. \eqref{e:bias_level}, 
	for any $t \in \re$,
	\begin{eqnarray}
	\Phi( t \!-\! \delta_R \sqrt{2 C_{\max}} )-p \leq 
	\Pr\Big( \tfrac{\sqrt{n}(\hat{\theta}_i - \theta^*_i)}{\sigma (\hat{\Omega}\hat{\Sigma}\hat{\Omega}^{\top})_{i,i}^{1/2}} \leq t 
	\,\Big| \mathcal{C}_i \Big)
	\leq 
	\Phi( t \!+\! \delta_R \sqrt{2 C_{\max}} ) + p.
		\label{e:bound_cdf_deblasso}
	\end{eqnarray}

\label{lemma:bounds_cdf_standardized_deblasso}
\end{lemma}

The next corollaries, 
provide upper and lower bounds for the probability of machines to send to the fusion center non-support and support indices, respectively.
Recall that $p_j^m$ denotes the probability that machine $m$ sends index $j$. 

\begin{corollary}
Assume that Algorithm \ref{alg:count_votes} is run with threshold $\tau>0$. 
Then, under the assumptions of 
Theorem 
\ref{theorem:deblasso_asymp_dist}, with $n$ and $d$ sufficiently large, $\forall j\notin S$,  
\begin{align}
p_j^m \leq & 2 \Phi^c\left( \tau - \delta_R \sqrt{2 C_{\max}} \right) + c_1 d \cdot e^{-c_2 n/K} + \frac{c_3}{d^2},
	\label{eq:bound_pj_non_support}
\end{align}
where  $c_1$, $c_2$, $c_3$ depend only on $C_{\min}$ and $C_{\max}$.
\label{corollary:upper_bound_prob_nonsupport_index}
\end{corollary}

\begin{corollary}
Assume that Algorithm \ref{alg:count_votes} is run with threshold $\tau>0$. 
Assume that all conditions of Lemma \ref{lemma:bounds_cdf_standardized_deblasso} and assumption \ref{assump:thetamin} hold. Then, for all $i\in \mathcal S$, 
\begin{align}
p_i^m \geq & \Phi^c\left( \tau + \delta_R \sqrt{2 C_{\max}} - \sqrt{2 r \ln d} \right) - c_1 d \cdot e^{-c_2 n/K} - \frac{c_3}{d^2},
	\label{eq:lower_bound_pi_support}
\end{align}
where  $c_1$, $c_2$ and $c_3$ depend only on  $C_{\min}$ and $C_{\max}$.
\label{corollary:lower_bound_prob_support_index}
\end{corollary}


\subsection{Probabilities of the voting scheme}

We first state some inequalities, used to theoretically analyze Algorithm \ref{alg:count_votes}.
The following are standard Gaussian tail bounds \citep[p. 10]{lin2010probability}, 
\begin{align}
    \frac{x}{\sqrt{2\pi}(x^2 + 1)}e^{-x^2/2} \leq 1-\Phi(x) \leq 
    \frac{1}{\sqrt{2\pi}x}e^{-x^2/2}, \quad \forall x>0.
    \label{e:ineq_gauss_cdf}
\end{align}
The next lemma, proven in section \ref{sec:prob_sending_indices}, bounds the Gaussian quantile function. 
\begin{lemma}
For all $t \geq 24$,
\begin{align}
\sqrt{2 \ln\left( \frac{t}{8\sqrt{\ln(t/8)}} \right)} 
\leq \Phi^{-1}\left( 1 - \frac{1}{t} \right) 
\leq \sqrt{2 \ln t}.
\label{e:ineq_gauss_quantile_function}
\end{align}
\label{lemma:ineq_gauss_quantile_function}
\end{lemma}

We also use the following inequality for 
a binomial random variable 
$V\sim Bin(M,p)$ \cite[exercise 2.11]{boucheron2013concentration}. 
For any  $0<p\leq a<1$, define $F(a,p) = a \ln\left(\frac{p}a\right) + (1-a) \ln\left( \frac{1-p}{1-a}\right)$. Then, 
\begin{align}
    \Pr\left(V > M a \right) \leq \left\{\left(\frac{p}{a}\right)^a\left(\frac{1-p}{1-a}\right)^{1-a}\right\}^M =   e^{M \cdot F(a,p)}.
    \label{e:ineq_concentr_binomial}
\end{align}
Since $\ln(1+x)\leq x$ for all $x\geq 0$, then  for $p \leq a$,
\begin{align}
F(a,p) 
&\leq a \ln\left(\frac{p}a\right) + a - p   
= a \ln\left(\frac{p}{a} e\right) - p .
	\label{e:ineq_concentr_binomial_exponent}
\end{align}


Recall that $V_j$ denotes the number of votes received by index $j$ at the fusion center (c.f. step 7 of Algorithm \ref{alg:count_votes}).
The following lemma shows that under suitable conditions, non-support indices receive a small total number of votes.


\begin{lemma} 
	Let $\delta \in (0,1)$. 
	Assume that $p^m_j\leq \delta$ for all non-support indices $j\not\in\mathcal{S}$ and all machines $m\in[M]$. 
	Assume that	$M$ and the threshold $V_T$ satisfy
	\begin{equation}
	V_T < 
	M < \frac{V_{T}}{e\delta} d^{-2/V_{T}}.
	\label{e:max_non_support_condition_M}
	\end{equation} 
	Then $\Pr\left(\max_{j\not\in S} V_j > V_{T}\right) \leq 1/d$.
			\label{lemma:max_votes_nonsupport}
\end{lemma}

Note that in a high dimensional low communication setting with $d\gg1$ and $p_j^m\leq \delta=c/d$,  
Eq. \eqref{e:max_non_support_condition_M} reads
$M< \frac{V_T}{ce}d^{1-2/V_T}$. Hence, if the voting threshold $V_T\geq 4$, then Lemma \ref{lemma:max_votes_nonsupport} holds for a wide range of possible values for $M$, up to $O(\sqrt{d})$. The next lemma considers the number of votes for support indices. 


\begin{lemma}
	Assume that $p^m_i \geq p_S$ for all support indices $i\in \mathcal S$ and all machines $m\in[M]$. 
	Assume that $M$, $p_S$ and the threshold $V_T$ satisfy
	\begin{equation}
				\label{e:min_M_support}
		M \geq \frac{2}{p_S}\left( \ln d + V_T \right).
	\end{equation}	
	Then $	\Pr\left(\min_{i\in \mathcal S} V_i \leq V_T\right) \leq \frac{K}d$.
\label{lemma:min_votes_support}
\end{lemma}


\section{Proofs of Theoretical Results}
\label{appendix:proofs_main_theorems}


\begin{proof}[Proof of Theorem \ref{theorem:support_recovery_large_tau}]
Recall that $\hat{\mathcal S}$ contains those indices that receive more than $V_T$ votes. Hence, for exact support recovery both $ \max_{j\not\in\mathcal{S}} V_j \leq V_T$
and $\min_{i\in\mathcal S} V_i > V_T$ must hold simultaneously. By a union bound, 
\begin{eqnarray}
	\Pr(\mathcal{\hat{S}} = \mathcal{S}) 
	&\geq &
1 - \Pr\left( \max_{j\not\in\mathcal{S}} V_j > V_T \right) - \Pr\left( \min_{i\in\mathcal{S}} V_i \leq 
V_T \right) .
\label{e:union_bound_maxVj_minVi_large_tau}
\end{eqnarray}
We now bound the two terms on the RHS of Eq. \eqref{e:union_bound_maxVj_minVi_large_tau}.
Eq. \eqref{eq:bound_pj_non_support} of 
Corollary \ref{corollary:upper_bound_prob_nonsupport_index}
provides an upper bound on $p_j^m$ for $j\notin \mathcal S$. Since $n$ satisfies \ref{assump:sample_size_support_recovery}, then for $n,d$ sufficiently large, 
the last two terms in \eqref{eq:bound_pj_non_support}
are at most $1/d$. 
Furthermore, by the definition of $\tau$ in Eq. 
\eqref{e:tau_large} and the condition on $\epsilon$ in
the theorem, 
\begin{align*}
	\tau - \delta_R \sqrt{2 C_{\max}} 
	=
	 \sqrt{2 \ln d} + \frac{\epsilon}{2}\sqrt{2 \ln d} - \delta_R \sqrt{2 C_{\max}} 
	 \geq \sqrt{2 \ln d}.
\end{align*}
Since $\Phi^c(\cdot)$ is a decreasing function, by the tail bound in Eq. \eqref{e:ineq_gauss_cdf},
\begin{align}
	2 \Phi^c\left( \tau - \delta_R \sqrt{2 C_{\max}} \right) 
	\leq 2 \Phi^c\left( \sqrt{2 \ln d} \right) \leq\frac{1}{\sqrt{\pi \ln d} \cdot d} 
	\leq \frac{1}{d}.
	\nonumber
\end{align}
Inserting this into \eqref{eq:bound_pj_non_support} yields $p_j^m \leq \frac{2}{d}$ for all $j\not\in\mathcal{S}$.

Next, the bound on the number of machines in Eq. \eqref{e:M_support_recovery_large_tau} with the threshold $V_T=\ln d$ imply that the condition \eqref{e:max_non_support_condition_M} of Lemma \ref{lemma:max_votes_nonsupport} holds with $\delta = \frac{2}{d}$. Indeed,   
\begin{align*}
	M  \leq \frac{V_T}{e \cdot \delta} d^{-2/V_T}
	= \frac{\ln d}{e \cdot (2/d)} d^{-2/\ln d} = \frac{\ln d}{2 e^3} \cdot d.
\end{align*}
Therefore, by Lemma \ref{lemma:max_votes_nonsupport}, 
\begin{equation}
	\Pr\left( \max_{j\not\in\mathcal{S}} V_j > \ln d \right) \leq \frac{1}{d} .
	\label{e:bound_prob_maxVj_large_tau}
\end{equation}

Next, let us examine support indices $i\in\mathcal{S}$. 
Eq. \eqref{eq:lower_bound_pi_support}
of Corollary \ref{corollary:lower_bound_prob_support_index},
provides a lower bound on $p_i^m$ for $i\in\mathcal S$. 
As above, for $n,d$ sufficiently large, the sum of the last two terms in \eqref{eq:lower_bound_pi_support} is at most $1/d$.
Also, 
by the definition of $\tau$ in Eq. \eqref{e:tau_large} and the condition on $\epsilon$,
$\tau + \delta_R \sqrt{2 C_{\max}} - \sqrt{2 r \ln d}
\leq (1 + \epsilon - \sqrt{r})\sqrt{2 \ln d}.
$
Hence, 
\begin{align}
p_i^m \geq & \Phi^c\left( (1 + \epsilon - \sqrt{r})\sqrt{2 \ln d} \right) - \frac{1}{d}, \quad \forall i\in\mathcal{S}.
\label{e:lower_bound_pi_large_tau}
\end{align}

Next, we wish to apply Lemma \ref{lemma:min_votes_support} with $V_T = \ln d$. 
The required condition, Eq.~\eqref{e:min_M_support} reads $M\geq 4 \ln d/p_S$. 
In light of Eq. \eqref{e:lower_bound_pi_large_tau}, a sufficient condition is that
\begin{align}
	p_i^m \geq \Phi^c\left( (1 + \epsilon - \sqrt{r})\sqrt{2 \ln d} \right) - \frac{1}{d} \geq \frac{4 \ln d}{M} = 
	p_S.
	\nonumber 
\end{align}
Eq. \eqref{e:M_support_recovery_large_tau} implies that $\frac{1}{d} \leq \frac{\ln d}{M}$.
Hence, for the above to hold, it suffices that
\begin{equation}
	\Phi^c\left( (1 + \epsilon - \sqrt{r})\sqrt{2 \ln d} \right) \geq \frac{5 \ln d}{M}.
	\label{e:condition2_pi_large_tau}
\end{equation}
For a high 
SNR $\sqrt{r}\geq 1+\epsilon$, Eq. \eqref{e:condition2_pi_large_tau}
is satisfied trivially. In this case, $1 + \epsilon - \sqrt{r}<0$. In addition, the condition 
$M\geq 10\ln d$
of Eq. \eqref{e:M_support_recovery_large_tau}
implies that $\Phi^c(0)=1/2\geq 5 \ln d/M$. 
In the low SNR case $\sqrt{r}<1+\epsilon$, applying
$\Phi^{-1}$ to Eq.~\eqref{e:condition2_pi_large_tau}, 
gives the requirement 
\begin{equation*}
	\sqrt{r} \geq 1 + \epsilon - \frac{1}{\sqrt{2 \ln d}} \Phi^{-1}\left( 1 - \frac{5 \ln d}{M} \right).
\end{equation*}
Since $M \geq 120\ln d$, we may apply Lemma \ref{lemma:ineq_gauss_quantile_function} with $t=\frac{M}{5 \ln d}$, which gives a lower bound on $\Phi^{-1}\left( 1 - \frac{5 \ln d}{M} \right)$.
Inserting this lower bound into the equation above yields the SNR condition in Eq. \eqref{e:SNR_support_recovery_large_tau} of the Theorem. 
Under this condition, 
Eq.~\eqref{e:min_M_support} holds
and we may invoke Lemma \ref{lemma:min_votes_support}
to obtain that 
\begin{align}
\Pr\left( \min_{i\in\mathcal{S}} V_i \leq \ln d \right) \leq \frac{K}{d}.
\label{e:bound_prob_minVi_large_tau}
\end{align}
Finally, Eq. \eqref{e:support_recovery_large_tau} follows by inserting Eqs. \eqref{e:bound_prob_maxVj_large_tau} and \eqref{e:bound_prob_minVi_large_tau} into \eqref{e:union_bound_maxVj_minVi_large_tau}.
\end{proof}


\begin{proof}[Proof of Theorem \ref{theorem:support_recovery_thresholds_not_fixed}]
As in the proof of Theorem \ref{theorem:support_recovery_large_tau}, $\Pr(\cal{\hat S} \neq \cal S)$ is lower bounded
by Eq.~\eqref{e:union_bound_maxVj_minVi_large_tau}.
We now bound the probabilities on the RHS of Eq. \eqref{e:union_bound_maxVj_minVi_large_tau}.
Let us start with the non-support indices $j\not\in\mathcal{S}$. 
Under Assumption 
\ref{assump:sample_size_support_recovery}, for $n,d$
sufficiently large and any $\alpha > 0$, 
the last two terms in 
Eq. \eqref{eq:bound_pj_non_support} of  
Corollary \ref{corollary:upper_bound_prob_nonsupport_index}
satisfy that 
\begin{align}
c_1 \cdot d e^{-c_2 n/K} + \frac{c_3}{d^2} \leq \frac{1}{d^{\alpha}}.
\label{e:upper_bound_pj_term2_thresholds_not_fixed}
\end{align}
In addition, by 
the definition of $\tau$ in Eq. \eqref{e:tau_not_fixed} 
and the assumption on $\epsilon$, 
\begin{align*}
\tau - \delta_R \sqrt{2 C_{\max}} 
=  
\sqrt{2 \alpha \ln d} + \frac{\epsilon}{2}\sqrt{2 \ln d} - \delta_R \sqrt{2 C_{\max}} \geq \sqrt{2 \alpha \ln d}.
\end{align*}
Hence, by the tail bound in Eq. \eqref{e:ineq_gauss_cdf} and the condition $\alpha \geq \frac{5}{\ln d}$,
\begin{align}
2 \Phi^c\left( \tau - \delta_R \sqrt{2 C_{\max}} \right) 
\leq 2 \Phi^c\left( \sqrt{2 \alpha \ln d} \right) \leq\frac{1}{\sqrt{ \alpha \pi \ln d} \cdot d^{\alpha}} 
\leq \frac{1}{d^{\alpha}} .
\label{e:upper_bound_pj_term1_thresholds_not_fixed}
\end{align}
Thus, inserting Eqs. \eqref{e:upper_bound_pj_term1_thresholds_not_fixed} and \eqref{e:upper_bound_pj_term2_thresholds_not_fixed} into \eqref{eq:bound_pj_non_support} yields $p_j^m \leq \frac{2}{d^{\alpha}}$ for all $j\not\in\mathcal{S}$.

Next, we apply Lemma \ref{lemma:max_votes_nonsupport} with $\delta = \frac{2}{d^{\alpha}}$. 
For $M=d^{\beta}$ and $d^{2/V_T} = e^{ (2\ln d)/V_T }$, the condition in Eq. \eqref{e:max_non_support_condition_M}, 
namely $e M\delta \cdot d^{2/V_T}< V_T < M$, reads as
\begin{align*}
2e \cdot d^{\beta - \alpha} e^{\frac{2 \ln d}{V_T}} < V_T < d^{\beta} ,
\end{align*}
which indeed holds by Eq. \eqref{e:alpha_VT_thresholds_not_fixed} of the Theorem.
Invoking Lemma \ref{lemma:max_votes_nonsupport} gives
\begin{align}
	\Pr\left( \max_{j\not\in\mathcal{S}} V_j > V_T \right) \leq \frac{1}{d}.
	\label{e:bound_prob_maxVj_thresholds_not_fixed}
\end{align}

Next, let us examine support indices $i\in\mathcal{S}$. 
A lower bound on $p_i^m$ is given by Eq. \eqref{eq:lower_bound_pi_support} of Corollary
\ref{corollary:lower_bound_prob_support_index}. 
By Eq. \eqref{e:tau_not_fixed} and the assumption on 
$\epsilon$, 
\begin{align*}
	\tau + \delta_R \sqrt{2 C_{\max}} - \sqrt{2 r \ln d}
	\leq (\sqrt{\alpha} + \epsilon - \sqrt{r})\sqrt{2 \ln d}.
\end{align*}
Hence, for $d,n$ sufficiently large, 
\begin{align}
	p_i^m \geq & \Phi^c\left( (\sqrt{\alpha} + \epsilon - \sqrt{r})\sqrt{2 \ln d} \right) - \frac{1}{d^{\alpha}}, \quad \text{for all } i\in\mathcal{S}.
	\label{e:lower_bound_pi_thresholds_not_fixed}
\end{align}
Next, we apply Lemma \ref{lemma:min_votes_support} with $p_S = \frac{2}{M}(\ln d + V_T) = \frac{2}{d^{\beta}}(\ln d + V_T)$,
for which Eq. \eqref{e:min_M_support} holds. 
To apply the lemma a necessary condition is that
$p_i^m\geq p_S$ for all $i\in\mathcal S$. 
By Eq. \eqref{e:lower_bound_pi_thresholds_not_fixed} and the definition of $p_S$, this condition holds if
\begin{align}
	\Phi^c\left( (\sqrt{\alpha} + \epsilon - \sqrt{r})\sqrt{2 \ln d} \right) 
	\geq \frac{2}{d^{\beta}}(\ln d + V_T) + \frac{1}{d^{\alpha}} .
	\label{e:lower_bound_Phi_complementary_thresholds_not_fixed}
\end{align}
Let us denote the RHS above by 
\begin{align*}
\tilde{p}_S = \frac{2}{d^{\beta}}(\ln d + V_T) + \frac{1}{d^{\alpha}}
= \frac{2(\ln d + V_T)d^{\alpha} + d^{\beta}}{d^{\alpha + \beta}}.
\end{align*}
Applying $\Phi^{-1}$ to Eq. \eqref{e:lower_bound_Phi_complementary_thresholds_not_fixed}
gives
\begin{align}
	\sqrt{r} \geq \sqrt{\alpha} + \epsilon - \frac{1}{\sqrt{2\ln d}} \Phi^{-1}(1 - \tilde{p}_S) .
	\label{e:lower_bound_sqrt_r_thresholds_not_fixed}
\end{align}
Next, we verify that $\tilde{p}_S < \frac{1}{24}$, so we may  apply Lemma \ref{lemma:ineq_gauss_quantile_function} to lower bound $\Phi^{-1}(1 - \tilde{p}_S)$. 
The condition $\alpha > \frac{5}{\ln d}$ implies that $1/d^\alpha< e^{-5} < 1/148$. 
Eq. \eqref{e:M_support_recovery_thresholds_not_fixed} implies that $d^\beta > 100 \ln d$ and 
thus $\frac{2\ln d}{ d^\beta}<1/50$. Similarly, by Eq. \eqref{e:alpha_VT_thresholds_not_fixed}, $\frac{2V_T}{d^\beta}\leq 1/50$. 
Hence, indeed $\tilde p_S \leq 2/50+e^{-5}<1/24$.

Applying Lemma \ref{lemma:ineq_gauss_quantile_function} with $t = 1/\tilde{p}_S$ gives
\begin{align*}
	\Phi^{-1}(1 - \tilde{p}_S) \geq \sqrt{2 \ln\left( \frac{1}{8 \tilde{p}_S \sqrt{\ln(1/8\tilde{p}_S)}} \right)} .
\end{align*}
Inserting this into Eq. \eqref{e:lower_bound_sqrt_r_thresholds_not_fixed}, 
a sufficient condition for $p_i^m \geq p_S$ for all $i\in\mathcal{S}$ is 
\begin{equation*}
\sqrt{r} \geq \sqrt{\alpha} + \epsilon - \sqrt{\frac{1}{\ln d} \ln\left( \frac{1}{8 \tilde{p}_S \sqrt{\ln(1/8\tilde{p}_S)}} \right)} .
\end{equation*}
By the definition of $\tilde{p}_S$, this inequality is exactly Eq. \eqref{e:SNR_support_recovery_thresholds_not_fixed}, which holds by assumption. Hence, $p_i^m \geq p_S$ for all $i\in\mathcal{S}$. Lemma \ref{lemma:min_votes_support} then implies that
\begin{align}
\Pr\left( \min_{i\in\mathcal{S}} V_i \leq \ln d \right) \leq \frac{K}{d}.
\label{e:bound_prob_minVi_thresholds_not_fixed}
\end{align}
Finally, Eq. \eqref{e:support_recovery_thresholds_not_fixed} follows by inserting Eqs. \eqref{e:bound_prob_maxVj_thresholds_not_fixed} and \eqref{e:bound_prob_minVi_thresholds_not_fixed} into \eqref{e:union_bound_maxVj_minVi_large_tau}.
\end{proof}



To prove Corollary \ref{corol:supprec_threshold_votes}, we 
use a result derived in the proof of 
Theorem A.1 of 
\citet{battey2018distributed} for distributed OLS. 
We state their result as the following auxiliary lemma
(see page 21 of their supplementary material).

\begin{lemma}
Consider the linear model in dimension $K$, 
\[
y = X^\top \beta^* + 
\sigma w,
\]
where $w\sim N(0,1)$, 
$X\sim N(0,\Sigma)$, and $\Sigma\in \mathbb{R}^{K\times K}$ satisfies
$0<C_{\min} \leq \sigma_{\min}(\Sigma)
\leq \sigma_{\max}(\Sigma)\leq C_{\max} < \infty$.
Assume that each of $M$ machines holds $n$ i.i.d. samples from this model, with $n>K$ and
let $N=M n$ be the total number of samples. 
Denote by $\hat \beta^m$ the least squares solution at the $m$-th machine
and $\hat\beta^{\mbox{\tiny LS}}$ the centralized least squares solution. 
If 
$M = O\left( \frac{N K}{(\max\{K,\ln N\})^2} \right)$, then 
\begin{align*}
    \Pr\left( \Big\|\tfrac1M\sum_m \hat\beta^m - \hat\beta^{\mbox{\tiny LS}}\Big\|_2 > C\tfrac{\sqrt{M}\max\{K,\ln N\}}{N} \right)
    \leq 
    c_1 M e^{-\max\{K, \ln N\}} + M e^{- c_2\frac{N}M},
\end{align*}
where $C,c_1,c_2>0$ are constants that do not depend on $K$ or $N$.
\label{lemma:Battey_etal_2018_theoA3}
\end{lemma}

\begin{proof}[Proof of Corollary \ref{corol:supprec_threshold_votes}]

We proceed similar to the proof of \citet[Cor. A.3]{battey2018distributed}. By the law of total probability, for any constant 
$T>0$, 
\begin{equation}
\Pr\left( \|\hat{\theta} - \hat\theta^{\mbox{\tiny LS}}\|_2 > T \right)
\leq \Pr\left( \left\{ \|\hat{\theta} - \hat\theta^{\mbox{\tiny LS}}\|_2 > T \right\}\cap\{\hat{\mathcal{S}}=\mathcal{S}\} \right) 
	+ \Pr(\hat{\mathcal{S}}\neq \mathcal{S}).  
\label{eq:theta_error}
\end{equation}
By Theorem \ref{theorem:support_recovery_large_tau}, 
the second term above is bounded by 
$\Pr(\hat{\mathcal{S}}\neq\mathcal{S}) \leq \frac{K+1}d$. When $\hat{\mathcal{S}}=\mathcal{S}$, $\hat{\theta}_j = \hat\theta^{\mbox{\tiny LS}}_j = 0$ for all $j\not\in\mathcal{S}$. Consequently, $\|\hat{\theta} - \hat\theta^{\mbox{\tiny LS}}\|_2 = \|\hat{\theta}_{\mathcal{S}} - \hat\theta^{\mbox{\tiny LS}}_{\mathcal{S}}\|_2$ and the first term on the RHS above is bounded by 
$
\Pr\left(  \|\hat{\theta}_{\cal S} - \hat\theta_{\cal S}^{\mbox{\tiny LS}}\|_2 >T  \right).
$
Applying Lemma \ref{lemma:Battey_etal_2018_theoA3}, with 
$T =  C\frac{\sqrt{M}\max\{K,\ln N\}}{N}$,
\[
\Pr\left(  \|\hat{\theta}_{\cal S} - \hat\theta_{\cal S}^{\mbox{\tiny LS}}\|_2 > C\frac{\sqrt{M}\max\{K,\ln N\}}{N}  \right)
    \leq
c_1 M e^{-\max\{K, \ln N\}} + M e^{- c_2\frac{N}M}.
\]
Since $e^{-\max\{K, \ln N\}} \leq e^{-\ln N}$ and $N=n \cdot M$, inserting this into \eqref{eq:theta_error}
gives  
$$
\Pr\left( \|\hat{\theta} - \hat\theta^{\mbox{\tiny LS}}\|_2 > C\frac{\sqrt{M} \max\{K,\ln N\}}{N} \right) \leq \frac{c_1}{n} + M e^{- c_2 \cdot n} +\frac{K + 1}{d}.
$$
Note that as $n,d\to\infty$, all probabilities on the right hand side above tend to zero. 
Hence, for  $M = O\left( \frac{N K}{(\max\{K,\ln N\})^2} \right)$ it follows that  $\|\hat{\theta} - \hat\theta^{\mbox{\tiny LS}}\|_2 = O_P\left(\sqrt{\frac{K}N}\right)$. Finally, since $\|\hat\theta^{\mbox{\tiny LS}} - \theta^*\|_2 = O_P\left(\sqrt{\frac{K}N}\right)$, then $\|\hat{\theta} - \theta^*\|_2 = O_P\left(\sqrt{\frac{K}N}\right)$ as well.
\end{proof}


\subsection{Minimax rate unattainable without condition 
on $\theta_{\min}$}
	\label{sec:minimax_rate}

In this section we show that without a lower bound on the nonzero coefficients of $\theta^*$, our method is {\em not} minimax rate optimal for mean squared error of estimation. 
We prove this claim under the simpler setting of distributed sparse normal means estimation. Let $\Theta[K] = \{\theta\in\mathbb{R}^d \,\big|\, \|\theta\|_0 \leq K \}$ denote the set of vectors with at most $K$ nonzero coefficients. 
Let $x_1,\ldots,x_N$ be  i.i.d. $N(\theta^*,\sigma^2 I)$, where $\theta^*\in  \Theta[K]$ and $\sigma>0$. 
In a centralized setting where all $N$ samples are at a single machine, 
by Proposition 8.20 of \cite{johnstone2019gaussianseq}, an upper bound on the minimax estimation risk is given by
\begin{equation}
	\inf_{\hat{\theta}} \sup_{\theta \in \Theta[K]} \mathbb{E}_\theta \| \hat{\theta} - \theta \|_2^2 \leq 2  \frac{\sigma^2}{N} K \left(1 + \ln d \right) .
	\label{eq:centralized_minmax}
\end{equation}
Moveover, this rate is achievable by hard thresholding the sample mean $\bar{x} = \sum_{i=1}^N x_i/N$ 
at threshold $\sigma \sqrt{2 (\ln d )/N}$ \citep[Thm. 8.21]{johnstone2019gaussianseq}.

Consider now a distributed setting with $M$ machines, each with $n = N/M$ samples, from a $K$-sparse
vector $\theta^*$ whose nonzero entries have the same value, 
\[
\theta^*_j = t_c \cdot (1-\delta) \quad \mbox{where } 0 < \delta < 1,\ \forall j\in \mathcal S.
\]
At each individual machine, its sample mean is a sufficient statistic. Hence, we may equivalently assume that each
machine has a single sample, but the noise level is
$\sigma/\sqrt{n}$.
Consider $t_c = \sigma \sqrt{2(\ln d) /n}$, which is approximately the maximal value of the noise coordinates. 
For $\delta> 0 $ and $d\gg 1$, the signal coordinates will be burried inside the noise. 
Specifically, 
there will be about $d^\gamma$ noise coordinates whose magnitude is larger than $t_c (1-\delta)$, where
$\gamma = \gamma(\delta)$ is monotonic increasing in $\delta$.
In this scenario, with a communication budget per machine $L\ll d^\gamma$, with high probability 
only noise coordinates will be sent to the fusion center. In general, each machine would send to the center a different random subset of indices. For a voting threshold $V_T$ chosen to suppress noise coordinates, 
no index would receive a sufficiently high number of votes. 
In this case, the fusion center would output an estimate $\hat \theta = 0$.  
The $L_2$-risk for this specific vector is 
\begin{equation}
	\mathbb{E}_{\theta^*}\|\hat \theta - \theta^*  \|^2_2 
	= K t_c^2 (1-\delta)^2 = 2 (1-\delta)^2  \frac{\sigma^2}{n} K \ln d 
	= 2 (1-\delta)^2 \frac{\sigma^2}{N} K(  \ln d ) \cdot M .
	\label{eq:risk_distributed} 
\end{equation} 
Comparing Eqs. \eqref{eq:risk_distributed} and \eqref{eq:centralized_minmax}, 
with $\delta$ such that $M (1-\delta)^2 > 1$, implies that our scheme is not minimax rate optimal. 
We conjecture that with a sub-linear communication budget, no scheme can achieve the centralized minimax rate.

\subsection{Comparison to Theoretical Results of \cite{amiraz2022distributed}}
	\label{sec:comparison_to_Amiraz}

\begin{proof}[SNR level in our Thm. \ref{theorem:support_recovery_large_tau} and in Thm. 2.C of \cite{amiraz2022distributed}]
\label{proof:comparison_Thm1_vs_Thm2C_Amiraz}
\hfill \break
\cite{amiraz2022distributed} consider  $M = d^{\beta} \ln d$ machines, where $0<\beta<1$, yielding 
an SNR requirement of $r \gtrsim (1 - \sqrt{\beta})^2$.
We show that with this number of machines, the lower bound in our Eq. \eqref{e:SNR_support_recovery_large_tau} 
results in a similar SNR requirement.
First, the term $c(d,M)$ may be written as
\begin{align*}
c(d,M) = 40\sqrt{\ln \left( \frac{d^{\beta} \ln d}{40 \ln d} \right)}
= 40\sqrt{ \beta\ln d - \ln 40 } .
\end{align*}
Next, consider the term inside the square root in Eq. \eqref{e:SNR_support_recovery_large_tau}. For $d\gg 1$, ${\ln c(d,M)}$ is negligible compared to $\beta\ln d$.  
Thus, 
\begin{align*}
\frac{1}{\ln d} \ln\left( \frac{d^{\beta} \ln d}{c(d,M) \cdot \ln d} \right)
= \frac{\beta \ln d - \ln c(d,M)}{\ln d} 
\approx \beta. 
\end{align*}
Inserting these into the lower bound in Eq. \eqref{e:SNR_support_recovery_large_tau} gives
\begin{align*}
r \geq \left( 1 + \epsilon - \sqrt{ \beta - \frac{\ln c(d,M)}{\ln d} } \right)^2 
\approx \left( 1 + \epsilon - \sqrt{ \beta } \right)^2 .
\end{align*}
This matches the SNR requirement in Theorem 2.C of \cite{amiraz2022distributed} whenever the bias term
$\epsilon$ 
due the debiased lasso is negligible.
\end{proof}


\begin{proof}[SNR level in our Thm. \ref{theorem:support_recovery_thresholds_not_fixed} and in Thm. 2.A of \cite{amiraz2022distributed}]
\label{proof:comparison_Thm2_vs_Thm2A_Amiraz}
\hfill \break
Theorem 2.A of \cite{amiraz2022distributed} considers a setting where $M\propto \ln d$ and $\tau = \sqrt{2 r \ln d}$.
Let us evaluate the SNR requirement in Eq. \eqref{e:SNR_support_recovery_thresholds_not_fixed} for a number of machines $M = c_1 \ln d$ and $\alpha = \frac{c_2}{\ln d}$, where $c_1 > 100$ and $c_2 \geq 5$ are constants. 
In this case, $d^{\alpha} = e^{c_2}$ and $\beta$ in Eq. \eqref{e:M_support_recovery_thresholds_not_fixed} satisfies $d^{\beta} = M = c_1 \ln d$.
Then, the condition in Eq. \eqref{e:alpha_VT_thresholds_not_fixed} can be written as
\begin{align*}
2e \frac{d^{\beta}}{d^{\alpha}} e^{\frac{2\ln d}{V_T}} < V_T \leq \frac{d^{\beta}}{100} - \ln d
\quad\Leftrightarrow\quad
2e \frac{c_1 \ln d}{e^{c_2}} e^{\frac{2\ln d}{V_T}} < V_T \leq \left( \frac{c_1 }{100} - 1 \right)\ln d.
\end{align*}
Thus, $V_T = c_3 \ln d$ for some $c_3 > 0$.

Let us examine the slowly varying term $c(\alpha, \beta, d, V_T)$ in \eqref{e:SNR_support_recovery_thresholds_not_fixed}. Notice
\begin{align*}
\frac{d^{\alpha + \beta}}{16(\ln d + V_T) d^{\alpha} + 8 d^{\beta} }
= \frac{ e^{c_2} \cdot c_1 \ln d }{ 16(\ln d + c_3 \ln d) e^{c_2} + 8 c_1 \ln d }
= \frac{ e^{c_2} \cdot c_1 }{ 16(1 + c_3 ) e^{c_2} + 8 c_1 }.
\end{align*}
Hence, $c(\alpha, \beta, d, V_T)$ is constant. Similarly, it follows that	
\begin{align*}
\ln \left( \frac{d^{\alpha + \beta}}{2(\ln d + V_T) d^{\alpha} + 8 d^{\beta} } \frac{1}{c(\alpha, \beta, d, V_T)} \right)
\end{align*}
is also constant, let us say, equal to $c_4 > 0$. Therefore, by Eq. \eqref{e:SNR_support_recovery_thresholds_not_fixed},
\begin{align*}
r \geq \left( \sqrt{\frac{c_2}{\ln d}} + \epsilon - \sqrt{\frac{c_4}{\ln d}} \right)^2
\geq \left( \frac{\sqrt{c_2} - \sqrt{c_4}}{\sqrt{\ln d}} + \epsilon \right)^2 .
\end{align*}
Hence, if $\epsilon \ll 1/\sqrt{\ln d}$, then $r \gtrsim \frac{1}{\ln d}$.
\end{proof}


\section{Proofs of Auxiliary Lemmas}
	\label{sec:proof_auxiliary_lemmas}


\subsection{Bounds on the lasso error}


\begin{proof}[Proof of Lemma \ref{lemma:prob_tildeB}]
	Consider the following event, for a scalar $\eta$ specified below,  
\[
\mathcal B_X = \left\{\vX\in\mathbb{R}^{n\times d}  \,\Big|\, \max_{1\leq i \leq d} \frac{\|X_i\|_2^2}n < 1+\eta \right\}.
\] 
Since the rows of $\vX$ are i.i.d. $\mathcal N(0,\Sigma)$, then for each column $i$, 
$\|X_i\|_2^2/\Sigma_{i,i}\sim \chi^2_n$.
Since $\Sigma_{i,i}\leq 1$, 
by standard concentration results for $\chi^2_n$ random variables and a union bound, 
for any $\eta \in[0,1]$,  
\[
\Pr(\mathcal B_X^c) \leq d e ^{-n \eta^2 / 8 }.
\]
Next, conditional on $\vX$, the vector $Z=\frac1n\vX^T W$ has entries $z_i\sim N(0,\frac{\sigma^2}n \frac{\|X_i\|_2^2}n)$. Hence, under the event $\mathcal B_X$, the random variable $\|Z\|_\infty$ is stochastically dominated by $\sigma \sqrt{\frac{1+\eta}n}\| \xi\|_\infty$, where $\xi \sim N(0,\vI_{n})$.
Therefore, 
\begin{eqnarray}
	\Pr\left(\tilde{\mathcal B}^c\right) 
	\leq 
	\Pr\left(\tilde{\mathcal B}^c \,|\, \mathcal B_X\right) + \Pr\left(\mathcal B_X^c\right) 
	\leq
	\Pr\left( \|\xi\|_\infty \geq \sqrt{\tfrac{16}{1+\eta} \ln d}\right) + de^{-n\eta^2/8} . \nonumber
\end{eqnarray}
By concentration results for Gaussian variables, the first term on the right-hand side is bounded by $2d \exp(-\frac{16}{2(1+\eta)}\ln d)$.  
Taking $\eta=1$ completes the proof. 
\end{proof}


The following condition is used to bound the estimation and prediction error of a lasso $\tilde{\theta}$. It holds under the restricted eigenvalue condition (c.f. Remark 3.3 in \cite{javanmard2018debiasing}).

\begin{definition}[Compatibility condition]
	A symmetric matrix $B\in\re^{d\times d}$ 
	satisfies the compatibility condition with respect 
	to a set $\mathcal{S}\subset[d]$, if for some $\phi>0$, and for all $\theta\in\re^d$ satisfying $\|\theta_{\mathcal{S}^c}\|_1 \leq 3\|\theta_{\mathcal{S}}\|_1$, 
	\begin{align}
		\|\theta_{\mathcal{S}}\|_1^2 \leq \frac{ |\mathcal{S}| \cdot \theta^{\top} B \theta}{\phi^2}.
			\label{eq:compatibility}
	\end{align}
	The compatibility constant $\phi^2(B,\mathcal S)$ is the largest scalar such that 
	\eqref{eq:compatibility} holds. 
	\label{def:compatitlity_condition}
\end{definition}


\begin{proof}[Proof of Lemma \ref{lemma:lasso_ell1Error_ell2predictionError}]
Since $\tilde{\theta}$ is a solution of Eq. \eqref{e:lasso_estimator}, 
it satisfies 
the inequality $\frac{1}{2n}\|Y-\vX\tilde{\theta}\|_2^2 + \lambda\|\tilde{\theta}\|_1 \leq \frac{1}{2n}\|Y-\vX\theta^*\|_2^2 + \lambda\|\theta^*\|_1$.
As in \citet[Lemma 6.1]{buhlmann2011statistics}, it implies that
\begin{align}
\frac{1}{2n}\| \vX (\tilde{\theta} - \theta^* ) \|_2^2 + \lambda \| \tilde{\theta} \|_1
\leq \frac{1}{n} W^{\top} \vX (\tilde{\theta} - \theta^*) + \lambda \|\theta^* \|_1.
\label{e:basic_inequality}
\end{align}
Next, under $\tilde{\mathcal{B}}(n,d,\sigma)$ with $\lambda \geq \sigma \sqrt{\frac{\ln d}{n}}$, it
follows that $\left\| n^{-1} W^{\top} \vX \right\|_{\infty} \leq \frac{\lambda}2$. 
Combining this with H\"{o}lder's inequality,
\begin{align*}
\left| \frac{1}{n} W^{\top} \vX (\tilde{\theta} - \theta^*) \right| 
\leq \left\| \frac{1}{n} W^{\top} \vX \right\|_{\infty} \|\tilde{\theta} - \theta^*\|_1  
\leq \frac{\lambda}{2} \|\tilde{\theta} - \theta^*\|_1.
\end{align*}
Plugging this inequality into Eq. \eqref{e:basic_inequality} gives,  
\begin{align}
\frac{1}{n}\| \vX (\tilde{\theta} - \theta^* ) \|_2^2 + 2\lambda \| \tilde{\theta} \|_1
\leq \lambda \|\tilde{\theta} - \theta^*\|_1 + 2 \lambda \|\theta^* \|_1 .
\label{e:basic_inequality_noNoise}
\end{align}
Let us rewrite this inequality into a more convenient form. Since $\theta^*_{\mathcal{S}^c} = 0$, then
\begin{align}
\|\tilde{\theta} - \theta^*\|_1 = \|\tilde{\theta}_{\mathcal{S}} - \theta^*_{\mathcal{S}}\|_1 + \| \tilde{\theta}_{\mathcal{S}^c}\|_1.
\label{e:decomp_theta_tilde_minus_theta_star}
\end{align}
Additionally, since $\|\tilde{\theta}\|_1 = \|(\tilde{\theta} - \theta^*)_{\mathcal{S}} + \theta^*_{\mathcal{S}}\|_1 + \|\tilde{\theta}_{\mathcal{S}^c}\|_1$, by the triangle inequality 
\begin{align}
\|\tilde{\theta}\|_1 \geq \|\theta^*_{\mathcal{S}}\|_1 - \|\tilde{\theta}_{\mathcal{S}} - \theta^*_{\mathcal{S}}\|_1 + \|\tilde{\theta}_{\mathcal{S}^c}\|_1 .
\label{e:ineq_theta_tilde_ell1}
\end{align}
Inserting \eqref{e:ineq_theta_tilde_ell1} 
and \eqref{e:decomp_theta_tilde_minus_theta_star}
into the LHS and RHS of Eq. \eqref{e:basic_inequality_noNoise} respectively, gives
\begin{align}
\frac{1}{n}\| \vX (\tilde{\theta} - \theta^* ) \|_2^2 + \lambda \| \tilde{\theta}_{\mathcal{S}^c} \|_1
\leq 3 \lambda \|\tilde{\theta}_{\mathcal{S}} - \theta^*_{\mathcal{S}}\|_1 .
\label{e:basic_inequality_beforeUseCompatibilityCondition}
\end{align}
To prove Eqs. \eqref{e:lasso_ell1Error} and \eqref{e:lasso_ell2predictionError}, notice that
\begin{align*}
\frac{1}{n}\| \vX (\tilde{\theta} - \theta^* ) \|_2^2 + \lambda \| \tilde{\theta} - \theta^* \|_1
= \frac{1}{n}\| \vX (\tilde{\theta} - \theta^* ) \|_2^2 + \lambda \| \tilde{\theta}_{\mathcal{S}^c} \|_1 + \lambda \| \tilde{\theta}_{\mathcal{S}} - \theta^*_{\mathcal{S}} \|_1.
\end{align*}
Combining this with Eq. \eqref{e:basic_inequality_beforeUseCompatibilityCondition} 
gives
\begin{align}
\frac{1}{n}\| \vX (\tilde{\theta} - \theta^* ) \|_2^2 + \lambda \| \tilde{\theta} - \theta^* \|_1
\leq 4 \lambda \|\tilde{\theta}_{\mathcal{S}} - \theta^*_{\mathcal{S}}\|_1 .
\label{e:basic_inequality_forCompatibilityCondition}
\end{align}
Since $\theta^*_{\mathcal{S}^c} = 0$, Eq. \eqref{e:basic_inequality_beforeUseCompatibilityCondition} also implies that $\tilde{\theta} - \theta^*$ satisfies the requirement in Definition \ref{def:compatitlity_condition} of the compatibility condition, i.e.,
\begin{align*}
\| \tilde{\theta}_{\mathcal{S}^c} - \theta^*_{\mathcal{S}^c} \|_1
= \| \tilde{\theta}_{\mathcal{S}^c} \|_1 
\leq 3 \|\tilde{\theta}_{\mathcal{S}} - \theta^*_{\mathcal{S}}\|_1.
\end{align*}
It follows from \citet[Remark 3.3]{javanmard2018debiasing} that under 
$\mathcal{B}_{\delta}(n,K,3)$, the compatibility constant satisfies $\phi^2(\hat{\Sigma}, \mathcal{S}) \geq (1-\delta)^2  C_{\min}$. Therefore
\begin{align*}
\|\tilde{\theta}_{\mathcal{S}} - \theta^*_{\mathcal{S}}\|_1^2 
\leq \frac{K \cdot (\tilde{\theta} - \theta^*)^{\top}\hat{\Sigma}(\tilde{\theta} - \theta^*)}{\phi^2(\hat{\Sigma}, \mathcal{S})}
\leq \frac{K \cdot (\tilde{\theta} - \theta^*)^{\top}\hat{\Sigma}(\tilde{\theta} - \theta^*)}{(1-\delta)^2 C_{\min}}.
\end{align*}
Since $\hat\Sigma=\frac1n\vX^\top\vX$ it follows that
$(\tilde{\theta} - \theta^*)^{\top}\hat{\Sigma}(\tilde{\theta} - \theta^*) =
\frac1n \|\vX(\tilde{\theta}-\theta^*)\|_2^2$. 
Therefore,
\begin{align*}
\|\tilde{\theta}_{\mathcal{S}} - \theta^*_{\mathcal{S}}\|_1^2
\leq \frac{K}{(1-\delta)^2 C_{\min}} \frac{1}{n}\| \vX (\tilde{\theta} - \theta^*) \|_2^2 .
\end{align*}
Inserting this inequality into Eq. \eqref{e:basic_inequality_forCompatibilityCondition}, gives
\begin{align}
\frac{1}{n}\| \vX (\tilde{\theta} - \theta^* ) \|_2^2 + \lambda \| \tilde{\theta} - \theta^* \|_1
\leq 4 \lambda \sqrt{\frac{K}{(1-\delta)^2 C_{\min}}} \sqrt{\frac{1}{n}\| \vX (\tilde{\theta} - \theta^*) \|_2^2} .
\label{e:basic_inequality_afterCompatibilityCondition}
\end{align}
Note that $n^{-1} \| \vX (\tilde{\theta} - \theta^*) \|_2^2$ appears on both sides of  \eqref{e:basic_inequality_afterCompatibilityCondition}. To have it only on the LHS, we apply the inequality
$
u v \leq \frac{u^2}{2\alpha} + \frac{\alpha v^2}{2}
$ {for all } $u,v\in\re$ and $\alpha>0$.
Choosing $u = \sqrt{\frac{\lambda^2 K}{(1-\delta)^2 C_{\min}}}$, $v=\sqrt{\frac{1}{n}\| \vX (\tilde{\theta} - \theta^*) \|_2^2}$ and $\alpha=1/4$, gives
\begin{align*}
4 \sqrt{\frac{\lambda^2 K}{(1-\delta)^2 C_{\min}}} \sqrt{\frac{1}{n}\| \vX (\tilde{\theta} - \theta^*) \|_2^2}
\leq 8 \frac{\lambda^2 K}{(1-\delta)^2 C_{\min}} + \frac{1}{2n}\| \vX (\tilde{\theta} - \theta^*) \|_2^2.
\end{align*}
Plugging this inequality into Eq. \eqref{e:basic_inequality_afterCompatibilityCondition} and rearranging terms, we get
\begin{align*}
\frac{1}{2n}\| \vX (\tilde{\theta} - \theta^* ) \|_2^2 + \lambda \| \tilde{\theta} - \theta^* \|_1
\leq 8 \frac{\lambda^2 K}{(1-\delta)^2 C_{\min}} .
\end{align*}
As both terms on the LHS above are non-negative, Eqs. \eqref{e:lasso_ell1Error} and \eqref{e:lasso_ell2predictionError} follow.
\end{proof}


\subsection{Proofs of lemmas about the precision estimator}

We start with some 
observations and auxiliary claims. 
By Assumption \ref{assump:gaussian_design_bounded_spectrum}, 
\begin{align}
	0 < \frac{1}{C_{\max}} \leq \Omega_{j,j} \leq \frac{1}{C_{\min}} < \infty, \quad \text{for all } j\in[d].
	\label{e:diagonal_values_Omega}
\end{align}
The next lemma bounds the norm of  the rows $\Omega_j$ of $\Omega$. 

\begin{lemma}
	Assume that \ref{assump:gaussian_design_bounded_spectrum} and \ref{assump:row-wise_sparsity} hold with row-wise sparsity $K_\Omega$. 
	Then, 
	\begin{align}
		\|\Omega_j\|_2 \leq \frac{1}{C_{\min}} \quad\text{and}\quad
		\|\Omega_j\|_1 \leq \frac{\sqrt{K_{\Omega} + 1}}{C_{\min}}.
		\label{e:bound_norm_row_precision}
	\end{align}
	\label{lemma:bound_norm_row_precision}
\end{lemma}
\begin{proof}
	Since $\Omega = \Omega^{\top}$, then $\Omega_j^{\top}\Omega_j = e_j^{\top} \Omega^2 e_j$.
	Hence, by Assumption \ref{assump:gaussian_design_bounded_spectrum}, 
	\begin{align*}
		\|\Omega_j\|_2
		= \sqrt{\Omega_j^{\top}\Omega_j} = \sqrt{e_j^{\top}\Omega^2 e_j} 
		\leq \sqrt{\sigma_{\max}(\Omega^2)} 
		=\frac{1}{\sigma_{\min}(\Sigma)}
		\leq \frac{1}{C_{\min}}.
	\end{align*}
	Next, by Assumption \ref{assump:row-wise_sparsity}, $\Omega_j$ has at most $K_{\Omega}+1$ non-zero entries. The second inequality in \eqref{e:bound_norm_row_precision} 
	follows from  
	Cauchy-Schwarz,  
	$\|\Omega_j\|_1 \leq \sqrt{K_{\Omega}+1} \cdot \| \Omega_j \|_2$.
\end{proof}


\begin{proof}[Proof of Lemma \ref{lemma:concentration_tilde_tau_j}]
	Let $r_j = X_j - \vX_{-j}\gamma_j$. Then, 
the first term of $\tilde{\tau}^2_j$ in Eq.~\eqref{e:tilde_tau} may
be written as 
\begin{equation}
\frac{1}{n}\|X_j - \vX_{-j}\tilde{\gamma}_{j}\|_2^2
= \frac1n \|r_j\|_2^2 + \frac{1}{n}\|\vX_{-j}(\gamma_{j} - \tilde{\gamma}_{j})\|_2^2 
+\frac2n r_j^\top \vX_{-j} (\gamma_j - \tilde \gamma_j).
	\label{e:decomp_Xj_minus_Xsimj_gamaj}
\end{equation}
Subtracting $1/\Omega_{j,j}$ from both sides of \eqref{e:tilde_tau}, 
by \eqref{e:decomp_Xj_minus_Xsimj_gamaj} and the triangle inequality,
\begin{align}
\begin{aligned}
\left| \tilde{\tau}_j^2 - \frac{1}{\Omega_{j,j}} \right|
\leq &\left| \frac{1}{n}\|r_j\|_2^2 - \frac{1}{\Omega_{j,j}} \right| + \frac{1}{n}\|\vX_{-j}(\gamma_{j} - \tilde{\gamma}_{j})\|_2^2 \\ 
&+ \frac{2}{n} \left| r_j^{\top} \vX_{-j}(\gamma_{j} - \tilde{\gamma}_{j}) \right| + \lambda_{\Omega} \|\tilde{\gamma}_{j}\|_1 .
\end{aligned}
\label{e:decomp_tauj_Omegajj}
\end{align}
We now bound each of the terms on the right hand side of the above equation. 
For the first term,  
by Eq. \eqref{e:model_rowwise_lasso}, $ r_j = X_j - \vX_{-j}\gamma_{j} = \varepsilon_j\sim N(0, \Omega_{j,j}^{-1}\vI_{n})$. Thus,
\begin{align}
\frac{1}{n}\|r_j\|_2^2
= \frac{1}{n}\|\varepsilon_j\|_2^2
\sim \frac{1}{n \Omega_{j,j}} \chi^2_{n}.
\label{e:norm_as_sum_chisquares}
\end{align}
By a standard concentration inequality for chi-squared variables \cite[p. 29]{wainwright2019high}, for $0 \leq t \leq 1/\Omega_{j,j}$, 
\begin{align*}
\Pr \left( \left| \frac{1}{n} \|\varepsilon_j\|_2^2- \frac{1}{\Omega_{j,j}} \right| \geq t \right)
= \Pr \left( \left| \frac{1}{n} \Omega_{j,j} \|\varepsilon_j\|_2^2  - 1 \right| \geq t \Omega_{j,j} \right)
\leq 2 e^{-n t^2 \Omega_{j,j}^2/8}.
\end{align*}
Let $t=\frac{1}{\Omega_{j,j}}\sqrt{24\frac{\ln d}n}$.
By the conditions of the lemma, $\Omega_{j,j}t\leq 1$ and thus, 
\begin{align}
\Pr \left( \left| \frac{1}{n}   \|\varepsilon_j\|_2^2 - \frac{1}{\Omega_{j,j}} \right| \geq \frac{1}{\Omega_{j,j}}\sqrt{24\frac{\ln d}{n}} \right)
\leq \frac{2}{d^3}.
	\label{eq:bound_norm_epsilon}
\end{align}
By Eq. \eqref{e:diagonal_values_Omega}, $\frac{1}{\Omega_{j,j}}\leq C_{\max}$. 
Since by \ref{assump:regularization_precision_estimator},
$\lambda_{\Omega} > 8\sqrt{C_{\max} \frac{\ln d}{n}}$,
then $t \leq \sqrt{C_{\max}} \lambda_\Omega$. Hence, 
Eqs. \eqref{e:norm_as_sum_chisquares} and \eqref{eq:bound_norm_epsilon} imply that with probability at least $1 - 2/d^3$,
\begin{align}
\left| \frac{1}{n}\|r_j\|_2^2 - \frac{1}{\Omega_{j,j}} \right| 
\leq \sqrt{C_{\max}} \lambda_{\Omega} .
\label{e:bound_Xj_minus_Xsimj_gamaj}
\end{align}

To bound the second term on the RHS of Eq. \eqref{e:decomp_tauj_Omegajj}, 
we apply Lemma 
\ref{lemma:lasso_ell1Error_ell2predictionError} with design matrix $\vX_{-j}$,
response $X_j$, 
and vector $\gamma_j$, 
 which by assumption \ref{assump:row-wise_sparsity}
is $K_\Omega$-sparse.
Note that $\vX_{-j}$ has i.i.d. rows  $N(0,\Sigma_{-j})$, where $\Sigma_{-j}$ is the matrix $\Sigma$ with row and column $j$ removed. Since
$\Sigma$ satisfies \ref{assump:gaussian_design_bounded_spectrum}, so does
$\Sigma_{-j}$.
Also since the row-wise noise level $\sqrt{1/\Omega_{j,j}} \leq \sqrt{C_{\max}}$, 
then \ref{assump:regularization_precision_estimator}
implies that $\lambda_\Omega$ satisfies the lower bound required by the lemma.
 Hence, under $\mathcal B_j$, 
Eq. \eqref{e:lasso_ell2predictionError} implies that
\begin{align}
\frac{1}{n}\|\vX_{-j}(\gamma_{j} - \tilde{\gamma}_{j})\|_2^2 
\leq \frac{16 K_{\Omega} \lambda_{\Omega}^2}{(1 - \delta)^2 C_{\min}}  .
\label{e:bound_Xsimj_gamaj_minus_tildeGamaj}
\end{align}

For the third term on the RHS of Eq. \eqref{e:decomp_tauj_Omegajj}, by the Cauchy-Schwarz inequality,
\begin{align*}
\frac{2}{n} \left| r_j^{\top} \vX_{-j}(\gamma_{j} - \tilde{\gamma}_{j}) \right|
\leq 2 \sqrt{ \frac{1}{n}}
{ \left\|  r_j\right\|_2 } 
\sqrt{ \frac{1}{n}} {\left\| \vX_{-j}(\gamma_{j} - \tilde{\gamma}_{j}) \right\|_2 }.
\end{align*}
Conditional on $\mathcal B_j$, applying the bounds in Eqs. \eqref{e:bound_Xsimj_gamaj_minus_tildeGamaj} and \eqref{e:bound_Xj_minus_Xsimj_gamaj}, and $\Omega_{j,j}^{-1} \leq C_{\max}$, with probability at least $1 - 2/d^3$,
\begin{align}
\frac{2}{n} \left| r_j^{\top} \vX_{-j}(\gamma_{j} - \tilde{\gamma}_{j}) \right|
\leq 2\sqrt{ 
\left( C_{\max} + \sqrt{C_{\max}} \lambda_{\Omega} \right)
\tfrac{16 K_{\Omega} \lambda_{\Omega}^2}{(1 - \delta)^2 C_{\min}} 
} .
\label{e:bound_innerprod_term_Xj_minus_Xsimj_gamaj}
\end{align}

Finally, we bound the last term in Eq. \eqref{e:decomp_tauj_Omegajj}. 
By the triangle inequality
\begin{align}
	\|\tilde{\gamma}_{j}\|_1 \leq \|\gamma_{j}\|_1 + \|\tilde{\gamma}_{j} - \gamma_{j}\|_1.
	\label{e:triangle_ineq_gamaj}
\end{align}
By Eq. \eqref{e:gamma_j}, $\|\gamma_j\|_1 = \|\Omega_{j}/\Omega_{j,j}\|_1 - 1$. 
Hence, Eqs. \eqref{e:diagonal_values_Omega} and \eqref{e:bound_norm_row_precision} imply that
\begin{align}
	\|\gamma_{j}\|_1 \leq \left\| \frac{\Omega_j}{\Omega_{j,j}} \right\|_1
	\leq C_{\max} \left\| \Omega_j \right\|_1
	\leq \frac{C_{\max}}{C_{\min}} \sqrt{K_{\Omega} + 1}.
	\label{e:ell1_gama}
\end{align}
Under the event $\mathcal{B}_j$, Lemma \ref{lemma:lasso_ell1Error_ell2predictionError} implies that
\begin{align}
	\|\tilde{\gamma}_{j} - \gamma_{j}\|_1 \leq \frac{8 K_{\Omega}}{(1 - \delta)^2 C_{\min}} \lambda_{\Omega}.
	\label{e:ell1_tildeGama_minus_gama}
\end{align}
Combining Eqs. \eqref{e:triangle_ineq_gamaj}, \eqref{e:ell1_gama} and \eqref{e:ell1_tildeGama_minus_gama} gives
\begin{align}
	\lambda_{\Omega} \|\tilde{\gamma}_{j}\|_1 \leq  \frac{C_{\max}}{C_{\min}} \sqrt{K_{\Omega} + 1} \cdot \lambda_{\Omega} + \frac{8 K_{\Omega} \lambda_{\Omega}^2}{(1 - \delta)^2 C_{\min}}  .
	\label{e:bound_ell1_tildeGama_minus_gama}
\end{align}
Hence, it follows from Eq. \eqref{e:decomp_tauj_Omegajj} and the bounds in Eqs. \eqref{e:bound_Xj_minus_Xsimj_gamaj}, \eqref{e:bound_Xsimj_gamaj_minus_tildeGamaj}, \eqref{e:bound_innerprod_term_Xj_minus_Xsimj_gamaj} and \eqref{e:bound_ell1_tildeGama_minus_gama} that, with probability at least $1-2/d^3$,
\begin{align*}
&\left| \tilde{\tau}_j^2 - \frac{1}{\Omega_{j,j}} \right|
\leq \sqrt{ C_{\max}} \lambda_{\Omega} 
+ \frac{ 24 K_{\Omega} \lambda_{\Omega}^2 }{(1 - \delta)^2 C_{\min}} \\
&\qquad + 2\sqrt{ 
\left( C_{\max} + \sqrt{C_{\max}}\lambda_{\Omega} \right)
\frac{16 K_{\Omega} \lambda_{\Omega}^2 }{(1 - \delta)^2 C_{\min}}
} 
+ \frac{C_{\max}}{C_{\min}} \sqrt{K_{\Omega} + 1} \cdot \lambda_{\Omega} \\
&\quad\leq \sqrt{K_{\Omega} + 1} \cdot \lambda_{\Omega} \Big\{  \sqrt{ C_{\max}}
+ \frac{ 24 \sqrt{K_{\Omega}} \lambda_{\Omega} }{(1 - \delta)^2 C_{\min}}
+ 8\sqrt{ 
\frac{ C_{\max} + \sqrt{C_{\max}}\lambda_{\Omega} }{(1 - \delta)^2 C_{\min}}
} 
+ \frac{C_{\max}}{C_{\min}} \Big\}.
\end{align*}
Since by assumption~\ref{assump:regularization_precision_estimator}, 
$\sqrt{K_{\Omega} + 1} \lambda_{\Omega} = \kappa_{\Omega} \sqrt{\frac{(K_{\Omega} + 1) \ln d}{n}} \leq 1$, 
the term inside the curly brackets above may be bounded
by some suitable constant $c=c(\delta,C_{\min},C_{\max})$
and the lemma follows. 
\end{proof}


\begin{proof}[Proof of Corollary \ref{corollary:bound_inverse_tilde_tau_j}]
By Lemma \ref{lemma:concentration_tilde_tau_j}, under $\mathcal B_j$,  with probability at least $1-2/d^3$,
\begin{align*}
\tilde{\tau}_j^2 \geq \frac{1}{\Omega_{j,j}} - c  \lambda_{\Omega}\sqrt{K_{\Omega}+1}.
\end{align*}
By Eq. \eqref{e:diagonal_values_Omega}, $1/\Omega_{j,j}\geq C_{\min}$. Also, since $\lambda_\Omega=  \kappa_{\Omega} \sqrt{\frac{\ln d}{n}}$, 
the condition $n\geq 4 c^2 \kappa_{\Omega}^2 (K_{\Omega}+1) \ln d /C_{\min}^2$ implies that $c\lambda_\Omega\sqrt{K_\Omega+1}\leq C_{\min}/2$.
Hence, $\tilde{\tau}_j^2 \geq C_{\min}/2$, and Eq. (\ref{eq:lower_bound_tilde_tau}) of the corollary follows. 
\end{proof}


\begin{proof}[Proof of Lemma \ref{lemma:bound_ell1_ell2_error_Omegahat_j}]
By Eq. \eqref{e:lasso_row_estimator}, $\hat\Omega_j = \tilde{\vC}_j/\tilde{\tau}_j^2$. 
Recall that $1/\tilde{\tau}^2_j$ is close to $\Omega_{j,j}$, and for any $p\in\{1,2\}$,  
$\left\| \tilde{\vC}_j - \frac{\Omega_j}{\Omega_{j,j}} \right\|_p = \left\| \tilde{\gamma}_j - \gamma_j \right\|_p$.
Hence, 
\begin{align}
\|\hat{\Omega}_j - \Omega_j\|_p = \Big\| \tfrac{1}{\tilde{\tau}^2_j}\tilde{\vC}_j - \Omega_j \Big\|_p
&= \left\| \tfrac{1}{\tilde{\tau}^2_j} \left( \tilde{\vC}_j - \tfrac{\Omega_j}{\Omega_{j,j}} \right) + \left( \tfrac{1}{\tilde{\tau}^2_j} - \Omega_{j,j} \right) \tfrac{\Omega_j}{\Omega_{j,j}} \right\|_p \nonumber \\
& \leq 
	\frac{1}{\tilde\tau_j^2}
	\left\{
\|\tilde{\gamma}_j - \gamma_j \|_p + 
|\tilde\tau_j^2 -\tfrac1{\Omega_{j,j}}
|\! \cdot \!
\| \Omega_j\|_p
\right\}.
	\nonumber
\end{align}
By Lemma \ref{lemma:concentration_tilde_tau_j} and Corollary \ref{corollary:bound_inverse_tilde_tau_j}, conditional on $\mathcal B_j$, 
with probability at least $1-2/d^3$, both $\frac{1}{\tilde\tau^2_j}\leq \frac2{C_{\min}}$ 
and $\left| \tilde{\tau}_j^2 - \Omega_{j,j}^{-1} \right| \leq c \sqrt{K_{\Omega} + 1} \cdot \lambda_{\Omega}$
hold. 
Thus, 
\begin{align}
\|\hat{\Omega}_j - \Omega_j\|_p 
\leq \frac2{C_{\min}} \left\{ \left\| \tilde{\gamma}_j - \gamma_j \right\|_p + c \sqrt{K_{\Omega} + 1} \cdot \lambda_{\Omega} \cdot \| \Omega_j\|_p \right\} .
	\nonumber
\end{align}
Next, we bound $\|\tilde{\gamma}_j-\gamma_j\|_p$.
Recall that $\tilde\gamma_j$ is the solution of a lasso with noise level $\sqrt{1/\Omega_{j,j}}\leq 
\sqrt{C_{\max}}$. Assumption \ref{assump:regularization_precision_estimator} allows us to invoke Lemma \ref{lemma:lasso_ell1Error_ell2predictionError}
and Corollary 
\ref{corollary:lasso_ell2Error}. These
imply that
under $\mathcal B_j$, $\left\| \tilde{\gamma}_j - \gamma_j \right\|_1 \leq \frac{8 K_{\Omega} \lambda_{\Omega}}{(1 - \delta)^2 C_{\min}}$
and
$\left\| \tilde{\gamma}_j - \gamma_j \right\|_2 \leq \frac{3 \sqrt{K_{\Omega}} \lambda_{\Omega}}{(1 - \delta)^2 C_{\min}}$. 
Finally, by Lemma \ref{lemma:bound_norm_row_precision}, 
$\|\Omega_j \|_2 \leq \frac1{ C_{\min}}$
and $\|\Omega_j \|_1 \leq \frac{\sqrt{K_\Omega + 1}}{C_{\min}}$. 
Inserting these bounds
into the equation above proves Eqs.~\eqref{e:bound_ell1_error_Omegahat_j} and \eqref{e:bound_ell2_error_Omegahat_j}.
\end{proof}



To prove Lemma \ref{lemma:bound_variance_debiased_lasso}
we shall use the following non-asymptotic bound on the error of the sample covariance $\hat{\Sigma}$, see \citet[Eq. (10)]{bickel2008covariance}.

\begin{lemma} Let 
	$\hat{\Sigma} = \frac{1}{n}\vX^{\top}\vX$, where $\vX \in \re^{n\times d}$ has i.i.d. rows from $N(0,\Sigma)$. If $\Sigma_{i,i} \leq 1 $ $\forall i\in[d]$, then for some universal constants
		$c_1,c_2$ and $t_{\max}$,  
	\begin{equation}
		\Pr\left( \max_{1\leq i,j\leq d} |\hat{\Sigma}_{i,j} - \Sigma_{i,j}| \geq t \right) \leq c_1 d^2 e^{-c_2 n t^2}, 
				\quad \forall\, 0\leq t \leq t_{\max}.
					\label{eq:BL_Bound} 
	\end{equation}
	\label{lemma:Eq10_BickelLevina2008}
\end{lemma}

\begin{proof}[Proof of Lemma \ref{lemma:bound_variance_debiased_lasso}]
By simple algebraic manipulations,
\[
\hat{\Omega}\hat{\Sigma}\hat{\Omega}^{\top}
=
\hat{\Omega}( \hat{\Sigma} -\Sigma )\hat{\Omega}^{\top} + ( \hat{\Omega} -\Omega )\Sigma ( \hat{\Omega} -\Omega )^{\top}
+ ( \hat{\Omega} -\Omega )\Sigma\Omega^{\top} + \Omega \Sigma ( \hat{\Omega} -\Omega )^{\top} + \Omega \Sigma \Omega^{\top}.
\]
Since $\Omega = \Sigma^{-1}$ and $\Omega = \Omega^{\top}$, subtracting $\Omega$ from both sides yields
\begin{align*}
\hat{\Omega}\hat{\Sigma}\hat{\Omega}^{\top} - \Omega
= \hat{\Omega}( \hat{\Sigma} -\Sigma )\hat{\Omega}^{\top} + ( \hat{\Omega} -\Omega )\Sigma ( \hat{\Omega} -\Omega )^{\top} + ( \hat{\Omega} -\Omega ) + ( \hat{\Omega} -\Omega )^{\top} .
\end{align*}
Applying the triangle inequality to the $j$-th diagonal element gives
\begin{equation}
	| (\hat{\Omega}\hat{\Sigma}\hat{\Omega}^{\top}- \Omega)_{j,j}  |
	\leq
	A_1 + A_2 + 2 A_3 
	\nonumber
\end{equation}
where the three terms are  
$A_1 = |( \hat{\Omega}( \hat{\Sigma} -\Sigma )\hat{\Omega}^{\top} )_{j,j}|,
A_2 = |( ( \hat{\Omega} -\Omega )\Sigma ( \hat{\Omega} -\Omega )^{\top} )_{j,j}|, 
$
and $
A_3 = | \hat{\Omega}_{j,j} - \Omega_{j,j}|.$
Then, for any $\gamma=\gamma_1+\gamma_2+2 \gamma_3$, 
\begin{eqnarray}
	\Pr\left(
	| (\hat{\Omega}\hat{\Sigma}\hat{\Omega}^{\top}- \Omega)_{j,j}  | \leq \gamma
	\right) & \geq & \Pr\left(A_1\leq \gamma_1 \cap
	A_2 \leq \gamma_2 \cap A_3\leq \gamma_3\right) \nonumber\\
	&\geq& \Pr(\mathcal {B}_j)
	\left(\sum_{i=1}^3\Pr(A_i\leq \gamma_i|\mathcal B_j ) -2 \right).
			\label{eq:Bj_A1_A2_A3}
\end{eqnarray}
We first lower bound $\Pr(\mathcal B_{j})$.
By its definition in Eq. \eqref{e:set_Btilde_delta}, 
\[
\Pr(\mathcal{B}_j) 
= \Pr(\mathcal{B}_{\delta,j} \cap \tilde{\mathcal{B}}_j )
\geq \Pr(\mathcal{B}_{\delta,j})
+ \Pr(\tilde{\mathcal{B}}_j ) -1.
\]
For the first term we apply Lemma \ref{lemma:RudelsonZhou2013_teo16} to $\vX_{-j}$, whose covariance matrix
$\Sigma_{-j}$ also satisfies assumption \ref{assump:gaussian_design_bounded_spectrum}. 
The second term $\Pr(\tilde{\mathcal B}_j)$ 
is bounded by Lemma~\ref{lemma:prob_tildeB}
applied to the noise vector $W=\vX_{-j}\gamma_j-X_j$ and the design matrix
$\vX_{-j}$. Choosing $\delta=1/\sqrt{8}$, these give,
\begin{equation}
		\Pr(\mathcal{B}_j) \geq 1- \frac2{d^3} - (d+2)e^{-n/8}.
			\nonumber
\end{equation}

Next, we bound the three terms $\Pr(A_i\leq \gamma_i|\mathcal B_j)$. For the first one, 
\[
	A_1 
	\leq
		\sum_{k,\ell} |\hat{\Omega}_{j,k}| |\hat{\Omega}_{j,\ell}| \big| \hat{\Sigma}_{k,\ell} -\Sigma_{k,\ell} \big| 
	 =  
	 \| \hat{\Sigma} -\Sigma \|_{\max} 
	  \|\hat{\Omega}_j\|_1^2
\]
where $\| \hat{\Sigma} -\Sigma \|_{\max} = \max_{k,j} | \hat{\Sigma}_{k,j} -\Sigma_{k,j} |$.
Hence, for any $q>0$, 
\begin{eqnarray}
\Pr\left(A_1 \leq  \gamma_1 |\mathcal B_j\right) 
 &\geq  &
\Pr\left(\|\hat\Sigma-\Sigma\|_{\max} \leq q
\cap \|\hat{\Omega}_{j}\|_1^2 \leq \tfrac{\gamma_1}q
 | \mathcal B_j\right)  \nonumber \\
& \geq & \Pr\left(  \|\hat\Sigma-\Sigma\|_{\max} \leq 
q | \mathcal B_j\right)  
      + \Pr\left( \|\hat{\Omega}_{j}\|_1^2 \leq \tfrac{\gamma_1}q  \big| \mathcal B_j \right)  -1. \nonumber
\end{eqnarray}
We bound the first term above as follows, 
\begin{eqnarray}
\Pr\left(  \|\hat\Sigma-\Sigma\|_{\max} \leq q | \mathcal B_j\right)  
	&= & 
\frac{\Pr(\|\hat\Sigma-\Sigma\|_{\max} \leq q \cap \mathcal B_j) }{ \Pr(\mathcal B_j)  }	\nonumber \\
	& \geq & \Pr(\|\hat\Sigma-\Sigma\|_{\max} \leq q) + \Pr(\mathcal B_j) -1. \nonumber
\end{eqnarray}
We choose $q=\frac{c\lambda_\Omega}{8\sqrt{C_{\max}}}$. Recall that by Assumption
\ref{assump:regularization_precision_estimator}, 
$\lambda_{\Omega} \geq 8\sqrt{C_{\max}}\sqrt{\frac{\ln d}{n}}$. Thus, 
$q \geq c \sqrt{\frac{\ln d}n}$. We set the constant $c$ sufficiently large, so that the probability in
\eqref{eq:BL_Bound} is at most $1/d^3$. 
Also note that by Assumption \ref{assump:sample_size_support_recovery}, we may 
assume sample size is sufficiently large
so $q<t_{\max}$. Then, by Lemma \ref{lemma:Eq10_BickelLevina2008}, 
\begin{align*}
	\Pr\left( \| \hat{\Sigma} -\Sigma \|_{\max} \leq \frac{ c \lambda_{\Omega} }{ 8\sqrt{C_{\max}}}  \right) \geq
	\Pr\left( \| \hat{\Sigma} -\Sigma \|_{\max} \leq c \sqrt{\frac{\ln d}{n}} \right) \geq 1-\frac{1}{d^3}.
\end{align*}

Next, we bound $\|\hat{\Omega}_j\|_1^2$. 
By Eq. (\ref{e:bound_norm_row_precision})
of Lemma \ref{lemma:bound_norm_row_precision}
and the triangle inequality, 
\begin{equation}
\|\hat{\Omega}_j\|_1 \leq \| \Omega_j\|_1+ \|\hat{\Omega}_j - \Omega_j\|_1
	\leq \frac{\sqrt{1+K_\Omega}}{C_{\min}} +  \|\hat{\Omega}_j - \Omega_j\|_1.
			\label{eq:bound_hat_Omega_i}
\end{equation}
Also, by Eq. \eqref{e:bound_ell1_error_Omegahat_j} of
Lemma \ref{lemma:bound_ell1_ell2_error_Omegahat_j}, 
with $\delta=1/\sqrt{8}$, 
\[
\Pr\left(\|\hat\Omega_j-\Omega_j\|_1 \leq c_1 (1+K_\Omega) \lambda_\Omega \,|\,\mathcal B_j\right) \geq 1-\frac2{d^3}.
\]
Combining this with \eqref{eq:bound_hat_Omega_i} and Assumption \ref{assump:regularization_precision_estimator}, whereby $\sqrt{1+K_\Omega}\lambda_\Omega \leq 1$,
gives 
\[
\Pr(\| \hat\Omega_j\|_1^2 \leq c (1+K_\Omega) \,|\,\mathcal B_j)
\geq 1-\frac2{d^3}.
\]
Hence, for $\gamma_1= q c(1+K_\Omega) = c_1 (1+K_\Omega) \lambda_\Omega$
where $c_1$ depends on $C_{\min},C_{\max}$, 
\begin{eqnarray}
\Pr(A_1\leq \gamma_1|\mathcal B_j)
& \geq &  
1 - \frac{5}{d^3} -  (d+2) e^{-n/8}.
	\label{eq:bound_A1} 
\end{eqnarray}

Next, we bound the second term $A_2$. By Assumption~\ref{assump:gaussian_design_bounded_spectrum},
\begin{align*}
A_2  
= |( \hat{\Omega}_i -\Omega_i )^{\top}\Sigma ( \hat{\Omega}_i -\Omega_i ) | 
\leq \sigma_{\max}(\Sigma) \| \hat{\Omega}_i -\Omega_i \|_2^2 
\leq C_{\max} \| \hat{\Omega}_i -\Omega_i \|_2^2 .
\end{align*}
By 
Eq. \eqref{e:bound_ell2_error_Omegahat_j} of
Lemma \ref{lemma:bound_ell1_ell2_error_Omegahat_j}, 
for $\delta = 1/\sqrt{8}$, combined with 
Assumption ~\ref{assump:regularization_precision_estimator}, $\lambda_\Omega \leq 1$, 
\begin{equation}
\Pr\left(A_2\leq c_2 (K_\Omega+1) \lambda_\Omega |\mathcal B_j\right) 
	\geq 1-\frac{2}{d^3}
	\label{eq:bound_A2}
\end{equation}
where $c_2$ depends on $C_{\min},C_{\max}$.

Finally, we bound the third term $A_3$. 
Recall from Eq.~\eqref{e:lasso_row_estimator} that $\hat{\Omega}_{j,j} = \tilde\tau_j^{-2}$, whereas by Eq. \eqref{e:diagonal_values_Omega}, $\Omega_{j,j} \leq 1/C_{\min}$. Hence, 
	\begin{align*}
		| \hat{\Omega}_{j,j} - \Omega_{j,j}| = \left| \frac{1}{\tilde\tau_j^2} - \Omega_{j,j} \right|
		= \frac{\Omega_{j,j}}{\tilde\tau_j^2} \left| \tilde\tau_j^2 - \Omega_{j,j}^{-1} \right|
		\leq 
		\frac{ 1 }{C_{\min} \tilde\tau_j^2} 
		\left| \tilde\tau_j^2 - \Omega_{j,j}^{-1} \right| .
	\end{align*}
Choosing $\gamma_3 = \frac{2}{C_{\min}^2} \cdot c(K_\Omega+1)\lambda_\Omega$, and employing 
Lemma \ref{lemma:concentration_tilde_tau_j}
and Corollary \ref{corollary:bound_inverse_tilde_tau_j}
\begin{eqnarray}
	\Pr(A_3\leq \gamma_3|\mathcal B_j) &\geq &
	 \Pr(\tfrac1{C_{\min}\tilde\tau_j^2} \leq \tfrac{2}{C_{\min}^2} |\mathcal B_j) +  \nonumber \\
	 & & 
	 \Pr(| \tilde\tau_j^2 - \Omega_{j,j}^{-1} | \leq c(K_\Omega+1) \lambda_\Omega  |\mathcal B_j) -1
	 \nonumber 	\\
	 & \geq & 1 - \frac{4}{d^3}.
	 	\label{eq:bound_A3} 
\end{eqnarray}	
Inserting \eqref{eq:bound_A1}, \eqref{eq:bound_A2}
and \eqref{eq:bound_A3} into \eqref{eq:Bj_A1_A2_A3} 
implies that with probability at
least $1  - \frac{13}{d^3} - 2(d+2)e^{-n/8}$,  
Eq. \eqref{e:bound_variance_debiased_lasso} holds. 
\end{proof}


\begin{proof}[Proof of Corollary \ref{corollary:bound_variance_debiased_lasso}]
	Recall that $1/C_{\max} \leq \Omega_{jj} \leq 1/C_{\min}$. Hence, 
	for (\ref{eq:bound_Omega_Sigma_Omega}) to hold, it suffices that
	$c(K_\Omega+1)\lambda_\Omega \leq \frac{1}{2C_{\max}}$.
	Plugging the relation $\lambda_\Omega = \kappa_\Omega \sqrt{\frac{\ln d}n}$ yields the required condition on sample size.
\end{proof}


\subsection{Properties of the debiased lasso estimator}
\label{sec:properties_debiased_lasso}

\begin{proof}[Proof of Lemma \ref{lemma:bounds_cdf_standardized_deblasso}]
Let $q_{i} = (\hat{\Omega}\hat{\Sigma}\hat{\Omega}^{\top})_{i,i}^{1/2}$. 
By Theorem \ref{theorem:deblasso_asymp_dist}, $\sqrt{n}(\hat{\theta}_i - \theta^*_i) = Z_i + R_i$, where $Z_i \big| \vX \sim N \left( 0, \sigma^2 q_{i}^2 \right)$ and $\Pr(|R_i| \geq \sigma \delta_R)\leq p/2$, with $\delta_R$ defined in Eq.~\eqref{e:bias_level} and $p$ defined in the Lemma.
Then, by the law of total probability
\begin{eqnarray}
\Pr\left( \frac{\sqrt{n}(\hat{\theta}_i - \theta^*_i)}{\sigma (\hat{\Omega}\hat{\Sigma}\hat{\Omega}^{\top})_{i,i}^{1/2}}
\leq t \,\Big|\, \mathcal C_i
\right)
&=&\! \Pr\left( \{ \tfrac{Z_i + R_i}{\sigma q_i} \leq t \} \cap \{|R_i|\leq \sigma \delta_R \} \,\Big|\, \mathcal{C}_i \right)  + \nonumber \\
& & \!\Pr\left( \{ \tfrac{Z_i + R_i}{\sigma q_i} \leq t \} \cap \{|R_i| > \sigma \delta_R \} \,\Big|\, \mathcal{C}_i \right)\!. \label{e:Zi_Ri_totalProb_RgtrDelta}
\end{eqnarray}
Let us upper bound the two terms on the RHS of \eqref{e:Zi_Ri_totalProb_RgtrDelta}. If $|R_i|\leq \sigma \delta_R$
then $ Z_i + R_i\geq Z_i - \sigma \delta_R$. Also, under $\mathcal C_i$ of Eq. \eqref{e:set_Ci}, 
$\frac{1}{q_{i}} \leq \sqrt{2 C_{\max}}$. 
Thus, 
\begin{eqnarray*}
	\Pr\left( \left\{ \tfrac{Z_i + R_i}{\sigma q_{i}} \leq  t \right\} \cap \{|R_i|\leq \sigma \delta_R \} \,\Big|\, \mathcal{C}_i \right)
	&\leq & \Pr\left( \tfrac{Z_i}{\sigma q_{i}} \leq \tfrac{\delta_R}{q_{i}} + t \,\Big|\, \mathcal{C}_i \right) \\
	& \leq & \Pr\left( \tfrac{Z_i}{\sigma q_{i}} \leq \delta_R \sqrt{2 C_{\max}} + t \,\Big|\, \mathcal{C}_i \right). 
\end{eqnarray*}
Since $\frac{Z_i}{\sigma q_i}|\vX$ is distributed as $N(0,1)$, the probability above equals $\Phi\left(\delta_R \sqrt{2 C_{\max}} + t \right)$, regardless of the conditioning on $\mathcal C_i$. 

Next, we bound the second term on the RHS of Eq. \eqref{e:Zi_Ri_totalProb_RgtrDelta}. Notice
\begin{align*}
\Pr\left( \{ Z_i + R_i \leq \sigma c_{i,i} t \} \cap \{|R_i| > \sigma \delta_R \} \,\Big|\, \mathcal{C}_i \right)
\leq \Pr\left( |R_i| > \sigma \delta_R \,\Big|\, \mathcal{C}_i \right).
\end{align*}
By Corollary \ref{corollary:bound_variance_debiased_lasso},
and Lemma \ref{lemma:bound_variance_debiased_lasso}, 
for sufficiently large $d$ and $n$,
 $\Pr\left( \mathcal{C}_i \right) \geq 1/2$. Thus
\begin{align*}
\Pr\left( |R_i| > \sigma \delta_R \,\Big|\, \mathcal{C}_i \right)
= \frac{\Pr\left( \{|R_i| > \sigma \delta_R\}\cap \mathcal{C}_i \right)}{\Pr\left( \mathcal{C}_i \right)}
	\leq
	2\Pr\left(|R_i| > \sigma \delta_R\right) \leq p.
\end{align*}
Inserting these two bounds into \eqref{e:Zi_Ri_totalProb_RgtrDelta} yields the right inequality in Eq. \eqref{e:bound_cdf_deblasso}.

To prove the left inequality in Eq. \eqref{e:bound_cdf_deblasso},  
we neglect the second term  
 in Eq.~\eqref{e:Zi_Ri_totalProb_RgtrDelta},  
and lower bound the first term. 
Under $\{|R_i|\leq \sigma \delta_R \}$, $Z_i + R_i \leq Z_i + \sigma \delta_R$. In addition, 
by 
$\Pr(A\cap B) \geq \Pr(A) + \Pr(B) - 1$,  
\begin{align}
&\Pr\left( \{ Z_i + R_i \leq \sigma q_{i} t \} \cap \{|R_i|\leq \sigma \delta_R \} \,\big|\, \mathcal{C}_i \right) 
	\nonumber \\
&\quad \geq \Pr\left( \{ Z_i + \sigma \delta_R \leq \sigma q_{i} t \} \cap \{|R_i|\leq \sigma \delta_R \} \,\big|\, \mathcal{C}_i \right) 
	\nonumber \\
&\quad \geq \Pr\left( \frac{Z_i}{\sigma q_i} + \frac{\delta_R}{q_i} \leq t \,\big|\, \mathcal{C}_i \right) 
+ \frac{\Pr\left( |R_i|\leq \sigma \delta_R \right)  - 1}{\Pr(\mathcal C_i)}.
\label{e:Zi_low_bound_unionBound}
\end{align}
For the first term on the RHS of Eq. \eqref{e:Zi_low_bound_unionBound}, since $\frac{1}{q_{i}} \leq \sqrt{2 C_{\max}}$,
\begin{align*}
\Pr\left( 
\tfrac{Z_i}{\sigma q_i} + \tfrac{\delta_R}{q_i} \leq t
\,\big|\, \mathcal{C}_i \right)
&\geq \Pr\left( \tfrac{Z_i}{\sigma q_{i}} \leq  t - \delta_R \sqrt{2 C_{\max}} \,\big|\, \mathcal{C}_i \right) 
= \Phi\left( t - \delta_R \sqrt{2 C_{\max}} \right) .
\end{align*}
Next, since $\Pr\left( \mathcal{C}_i \right) \geq 1/2$, we
bound the second term in Eq.~\eqref{e:Zi_low_bound_unionBound} as follows 
\begin{eqnarray*}
	\frac{\Pr\left( |R_i|\leq \sigma \delta_R \right)  - 1}{\Pr(\mathcal C_i)} = -\frac{\Pr(|R_i| >  \sigma \delta_R )}{\Pr(\mathcal C_i)} \geq  -p. 
\end{eqnarray*}
Combining the above proves the left inequality of Eq. \eqref{e:bound_cdf_deblasso}.
\end{proof}


\begin{proof}[Proof of Corollary \ref{corollary:upper_bound_prob_nonsupport_index}]
By definition, the probability of machine $m$ to send index $j$ to the fusion center is $p_j^m = \Pr(|\xi_j^m| > \tau)$ where $\xi$ is defined in \eqref{e:standardized_debiased_lasso}. 
We bound $p_j^m$ by conditioning on the event $\mathcal{C}_j$ of Eq. \eqref{e:set_Ci}. By the law of total probability,
\begin{align}
p_j^m 
\leq \Pr\left(\xi_j^m > \tau \big| \mathcal{C}_j \right)
+ \Pr\left(\xi_j^m <- \tau \big| \mathcal{C}_j \right)
+ \Pr( \mathcal{C}_j^c ). \nonumber
\end{align}
By the LHS of Eq. \eqref{e:bound_cdf_deblasso}, the first term on the RHS above 
is bounded by
\begin{align*}
	\Pr\left(\xi_j^m > \tau \big| \mathcal{C}_j \right) =
	1-\Pr(\xi\leq \tau|\mathcal C_j) \leq 
\Phi^c(\tau- \delta_R \sqrt{2 C_{\max}}) + p.
\end{align*}
By the RHS of Eq. \eqref{e:bound_cdf_deblasso}, the second term above is bounded as 
$$
\Pr\left(\xi_j^m <- \tau \big| \mathcal{C}_j \right) \leq \Phi(-\tau  + \delta_R\sqrt{2 C_{\max}}) + p
= \Phi^c(\tau  - \delta_R\sqrt{2 C_{\max}}) + p.
$$
Finally, 
by Corollary \ref{corollary:bound_variance_debiased_lasso}
and Lemma \ref{lemma:bound_variance_debiased_lasso},  
$
\Pr( \mathcal C_j^c) \leq  
13/d^3 + 2(d+2)e^{-n/8}$. 
Combining it all yields Eq. \eqref{eq:bound_pj_non_support}.
\end{proof}


\begin{proof}[Proof of Corollary \ref{corollary:lower_bound_prob_support_index}]
Without loss of generality, we assume that $\theta^*_i > 0$.
As in the previous proof, we condition on the event $\mathcal{C}_i$ defined in Eq. \eqref{e:set_Ci}. 
Then, 
\begin{equation}
	p_i^m 
	\geq 
	\Pr\left(|\xi_i^m| > \tau \big| \mathcal{C}_i \right) \Pr(\mathcal{C}_i) 
	\geq 
	\Pr\left(\xi_i^m > \tau \big| \mathcal{C}_i \right) \Pr(\mathcal{C}_i). \nonumber
\end{equation}
Let $q_i = (\hat{\Omega}\hat{\Sigma}\hat{\Omega}^{\top})_{i,i}^{1/2}$. By the definition of $\xi_i^m$ in \eqref{e:standardized_debiased_lasso},
\begin{align*}
\Pr\left(\xi_i^m \leq \tau \big| \mathcal{C}_i \right) 
= \Pr\left(\frac{ \sqrt{n} \hat{\theta}_i }{\sigma q_i} \leq \tau \Big| \mathcal{C}_i \right)
= \Pr\left(\frac{ \sqrt{n} ( \hat{\theta}_i - \theta_i^* ) }{\sigma q_i} \leq \tau - \frac{ \sqrt{n} \theta_i^* }{\sigma q_i} \Big| \mathcal{C}_i \right) .
\end{align*}
Under $\mathcal{C}_i$, $\frac{1}{q_i} \geq \sqrt{\frac{C_{\min}}{2}}$, whereas by  assumption \ref{assump:thetamin}, 
$\theta_i^*\geq \theta_{\min}$. Given 
the definition of the SNR $r$ in 
\eqref{e:thetamin}, 
\[
\tau - \frac{ \sqrt{n} \theta_i^* }{\sigma q_i} 
\leq \tau - \frac1\sigma \sqrt{\frac{n C_{\min}}2}\theta_{\min}
= \tau - \sqrt{2 r \ln d}.
\]
By the RHS of Eq.~\eqref{e:bound_cdf_deblasso} of Lemma \ref{lemma:bounds_cdf_standardized_deblasso},
\begin{eqnarray*}
	\Pr\left(\xi_i^m \leq \tau \big| \mathcal{C}_i \right) & \leq &
	\Pr\left(\frac{ \sqrt{n} ( \hat{\theta}_i - \theta_i^* ) }{\sigma c_{i,i}} \leq \tau - \sqrt{2 r \ln d} \,\Big|\, \mathcal{C}_i \right)  
	\\
	& \leq &
	\Phi\left( \tau - \sqrt{2 r \ln d} + \delta_R \sqrt{2 C_{\max}} \right) 
	+ p.
\end{eqnarray*}
Finally, by 
Corollary \ref{corollary:bound_variance_debiased_lasso}
and Lemma \ref{lemma:bound_variance_debiased_lasso}, 
$\Pr(\mathcal C_i) \geq 1- 13/d^3 - 2(d+2)e^{-n/8}$. 
Hence, 
\[
p_i^m \geq (\Phi^c\left( \tau - \sqrt{2 r \ln d} + \delta_R \sqrt{2 C_{\max}} \right) - p ) \cdot (1- 13/d^3 - 2(d+2)e^{-n/8}),
\]
from which Eq. \eqref{eq:lower_bound_pi_support} of the corollary follows.\end{proof}


\subsection{Probabilities of sending support or non-support indices}
\label{sec:prob_sending_indices}


\begin{proof}[Proof of Lemma \ref{lemma:ineq_gauss_quantile_function}]
\label{proof:lemma_ineq_gauss_quantile_function}
	Plugging $x=\sqrt{2\ln t}$ into 
	Eq. \eqref{e:ineq_gauss_cdf}   gives 
$$
\Phi^{c}(\sqrt{2 \ln t}) \leq \frac{1}{\sqrt{2\pi}}\frac1t \frac{1}{\sqrt{2\ln t}}.
$$
For 
$t\geq \exp(1/4\pi)\approx 1.08$ the right inequality
of Eq. \eqref{e:ineq_gauss_quantile_function} holds. 

For the left inequality of Eq. \eqref{e:ineq_gauss_quantile_function}, we look for an $x$ such that $\Phi^{c}(x) \geq 1/t$. 
Since $\frac{1+x^2}x \leq 2x$ for all $x\geq 1$, 
the lower bound of Eq. \eqref{e:ineq_gauss_cdf} reads
\[
\Phi^c(x) \geq \frac{1}{2x}  \frac{e^{-x^2 / 2}}{\sqrt{2\pi}}.
\]
Changing variables $x=\sqrt{2 \ln (y / \sqrt{\ln y})}$,
 $\Phi^{c}(x) \geq 1/t$ if the following holds, 
\[
\sqrt{\frac{ \ln y}{\ln y - \frac12\ln \ln y  }  } \cdot  \frac{1}{4\sqrt{\pi}} \cdot \frac1y \geq \frac1t.
\]
Since $\sqrt{\frac{\ln y}{\ln y - \frac{1}{2}\ln \ln y}} \geq \frac{\sqrt{\pi}}{2} $ for all $y \geq 3$, choosing
$y = t/8$ satisfies the required inequality. Hence
for $t\geq 24$, the LHS of Eq. \eqref{e:ineq_gauss_quantile_function}
holds as well.
\end{proof}


\begin{proof}[Proof of Lemma \ref{lemma:max_votes_nonsupport}]
	For any index $j\notin S$, $V_j$ is the sum of $M$ independent Bernoulli random variables with probabilities $p_j^m\leq \delta$. Thus, $V_j$ is stochastically dominated by $V\sim Bin(M,\delta)$. 
	By a union bound, 
	\[
	\Pr\left( \max_{j\not\in S} V_j > V_{T}  \right) \leq d \Pr\left(V > V_{T}\right) .
	\]
	Eq. \eqref{e:max_non_support_condition_M} implies that $V_{T} > M\delta$. Thus, by Eq. \eqref{e:ineq_concentr_binomial} 
	with $a=V_{T}/M$ and $p=\delta$,
	\[
	\Pr\left( \max_{j\not\in S} V_j > V_{T}  \right) \leq
	d \exp\left(V_{T} \ln\left( \frac{\delta e}{V_T / M} \right) - M\delta\right). 
	\]
	By Eq. \eqref{e:max_non_support_condition_M}, 
	$V_T \ln(\frac{\delta e M}{V_T}) \leq -2 \ln d$. 
	Hence, the above is bounded by $1/d$.
\end{proof}


To prove Lemma \ref{lemma:min_votes_support}, we use Chernoff's bound  
\citep{hagerup1990guided}.

\begin{lemma}
	\label{lemma:Chernoff}
	Suppose $X_1,\ldots X_n$ are independent Bernoulli random variables, and let $S$ denote their sum. 
	Then for any $\epsilon \in[0,1]$, 
	\begin{equation}
		\label{e:Chernoff}
		\Pr\left(S \leq (1-\epsilon) \mathbb{E}[S]\right) \leq \exp\left(-\epsilon^2 \mathbb{E}[S]/2\right). 
	\end{equation}
\end{lemma}

\begin{proof}[Proof of Lemma \ref{lemma:min_votes_support}]
	For any index $i\in \mathcal S$, $V_i$ is the sum of $M$ independent Bernoulli random variables with probabilities $p_i^m\geq p_S$. Hence, $V_i$ stochastically dominates $V\sim Bin(M,p_S)$. Since $|\mathcal S| = K$, then by a union bound, 
	\[
	\Pr\left( \min_{i\in \cal S} V_i \leq V_{T}  \right) \leq K \Pr\left( V \leq V_{T}\right).
	\]
	Using Chernoff's bound \eqref{e:Chernoff} with 
	$\epsilon = 1- V_T/(M p_S)$,
	\begin{equation*}
		\Pr\left( \min_{i\in \cal S} V_i \leq V_{T}  \right)
		\leq
		K \exp\left(-(1-\tfrac{V_T}{M p_S})^2   \tfrac{M p_S}2\right)  
		\leq
		K \exp\left(- \tfrac{M p_S}2\right) \cdot \exp(V_T) 
	\end{equation*}
	Eq. \eqref{e:min_M_support} implies that the RHS above
	is smaller than $K/d$. 
\end{proof}

\end{appendix}

\begin{acks}[Acknowledgments]
The authors would like to thank two anonymous referees for their constructive comments that improved the quality of this paper.
\end{acks}


\bibliographystyle{imsart-nameyear} 
\bibliography{bibfile}       


\end{document}